\documentclass[lettersize,journal]{IEEEtran}
\usepackage{amsmath,amsfonts}
\usepackage{algorithmic}
\usepackage{algorithm}
\usepackage{array}
\usepackage[caption=false,font=normalsize,labelfont=sf,textfont=sf]{subfig}
\usepackage{textcomp}
\usepackage{stfloats}
\usepackage{xurl}
\usepackage{verbatim}
\usepackage{graphicx}
\usepackage{cite}
\usepackage{orcidlink}
\usepackage{multirow}

\newcommand\bluecite[1]{\textcolor{blue}{\cite{#1}}}

\begin{document}
\title{Open TutorAI: An Open-source Platform for Personalized and Immersive Learning with Generative AI 
}

\author{
    Mohamed {EL HAJJI}$^{1,2,*}$\orcidlink{0000-0002-0327-8249}, Tarek {AIT BAHA}$^{1,4}$\orcidlink{0000-0001-6201-1985}, Aicha {DAKIR}$^{1,3}$\orcidlink{0009-0001-4303-0942}, Hammou {FADILI}$^{5}$\orcidlink{0000-0001-6283-3807}, Youssef {ES-SAADY}$^{1,3}$\orcidlink{0000-0002-4934-2322} \\
    
    \IEEEauthorblockA{$^{1}$IRF-SIC Laboratory, Ibnou Zohr University, Agadir 80000, Morocco} \\
    \IEEEauthorblockA{$^{2}$Regional Center for Education and Training Professions-Souss Massa, Agadir 80000, Morocco} \\
    \IEEEauthorblockA{$^{3}$Polydisciplinary Faculty of Taroudant, Ibnou Zohr University, Taroudant 83000, Morocco} \\
    \IEEEauthorblockA{$^{4}$Higher School of Technology of Guelmim,  Ibnou Zohr University, Guelmim 81000, Morocco} \\
    \IEEEauthorblockA{$^{5}$Paragraphe laboratory, Paris 8 and CY Cergy Paris Universities, Cergy, France} \\
    \thanks{$^{*}$ Corresponding author's email address: \\ m.elhajji@crmefsm.ac.ma (Mohamed EL HAJJI \orcidlink{0000-0002-0327-8249})}
}
\maketitle

\begin{abstract}
Recent advances in artificial intelligence have created new possibilities for making education more scalable, adaptive, and learner-centered. However, existing educational chatbot systems often lack contextual adaptability, real-time responsiveness, and pedagogical agility. which can limit learner engagement and diminish instructional effectiveness. Thus, there is a growing need for open, integrative platforms that combine AI and immersive technologies to support personalized, meaningful learning experiences. This paper presents \textbf{Open TutorAI}\footnote{\url{https://github.com/Open-TutorAi/open-tutor-ai-CE/}}, an open-source educational platform based on large language models (LLMs) and generative technologies that provides dynamic, personalized tutoring. The system integrates natural language processing with customizable 3D avatars to enable multimodal learner interaction. Through a structured onboarding process, it captures each learner’s goals and preferences in order to configure a learner-specific AI assistant. This assistant is accessible via both text-based and avatar-driven interfaces. The platform includes tools for organizing content, providing embedded feedback, and offering dedicated interfaces for learners, educators, and parents. This work focuses on learner-facing components, delivering a tool for adaptive support that responds to individual learner profiles without requiring technical expertise. Its assistant-generation pipeline and avatar integration enhance engagement and emotional presence, creating a more humanized, immersive learning environment. Embedded learning analytics support self-regulated learning by tracking engagement patterns and generating actionable feedback. The result is Open TutorAI, which unites modular architecture, generative AI, and learner analytics within an open-source framework. It contributes to the development of next-generation intelligent tutoring systems. It provides a flexible and extensible foundation for advancing personalized and inclusive education.

\end{abstract}


\section{Introduction}
Educational technologies such as Learning Management Systems (LMS) and Massive Open Online Courses (MOOCs) have had a considerable impact on the processes of teaching and learning, enabling the adoption of innovative pedagogical practices \bluecite{ferguson2014innovative}. In recent years, significant advancements in the domains of Artificial Intelligence (AI) have had a profound impact on the educational ecosystem. These developments have enabled novel forms of interaction with learning content and have provided instructors with valuable support in addressing the diverse learning needs of their students \bluecite{spriggs2025personalizing}. These developments offer the potential for personalized, scalable, and adaptive learning experiences. Nonetheless, while these technologies play an increasingly important role, they have not fully addressed certain core pedagogical responsibilities traditionally carried out by human educators, such as maintaining student motivation, delivering individualized feedback, and dynamically adapting instructional strategies. Conventional edu-chatbot systems frequently lack real-time interactivity and context-aware personalization, which can lead to reduced student engagement and suboptimal learning outcomes \bluecite{islam2015learning,al2021advantages}.

To overcome these limitations, researchers have turned to more adaptive, AI-powered systems that better emulate human educational support. Among the most promising AI-based tools in education are Intelligent Tutoring Systems (ITS), which aim to replicate aspects of human guidance by adapting to individual learners’ needs, tracking their progress, and providing context-sensitive support \bluecite{mousavinasab2021intelligent}. A complementary trend is the integration of conversational agents or educational chatbots, which have been employed to assist learners by transmitting knowledge, facilitating access to information, and automating routine tasks. For instance, voice-based digital assistants have been developed to facilitate home learning environments \bluecite{tarek2022towards}. In this capacity, chatbots facilitate the learning process by enhancing information accessibility and addressing repetitive queries, such as providing responses to frequently asked questions about examination dates or office hours \bluecite{sandoval2018design}.

Beyond these practical functions, some chatbots go further by adopting a monitoring role that supports learners’ self-regulation and reflective learning. In such models, the dialogue centers on the learner’s development, aiming to encourage planning, analysis, and self-assessment. For instance, the chatbot developed by \bluecite{gabrielli2020chatbot} offers life-skills tutoring to adolescents, with a specific emphasis on mental health support and self-reflection. These types of conversational agents demonstrate the expanding pedagogical potential of chatbots in supporting holistic learner development. These examples highlight the evolving capabilities of chatbots in education, from handling logistical queries to promoting reflective learning and emotional support. However, despite these advancements, many existing educational chatbot systems remain limited in their adaptability, often relying on scripted or text-based interactions with minimal personalization. Furthermore, they typically do not incorporate the latest developments in generative AI, and their design is frequently constrained by closed, proprietary architectures that hinder customization and scalability from the educator’s perspective.

To address these challenges, recent breakthroughs in LLMs present new opportunities to overcome these limitations \bluecite{brown2020language}. LLMs possess advanced capabilities in natural language understanding and generation, allowing for the development of dynamic AI assistants that engage in real-time dialogue, provide contextualized feedback, and tailor learning paths to individual needs \bluecite{sharma2025role}. However, despite their potential, many current LLM-based educational platforms fall short in terms of flexibility, ethical deployment, and smooth integration with institutional LMS infrastructures. Moreover, their implementation often requires significant technical expertise and raises important concerns about data privacy, transparency of AI reasoning, and alignment with sound pedagogical principles.

Building on the increasing role of AI-driven personalization in education, the emergence of the metaverse and immersive 3D digital environment are further transforming how users interact, collaborate, and learn online \bluecite{yeganeh2025future}. Central to these virtual spaces are avatars, which serve as key elements of self-representation and social presence \bluecite{hepperle2022aspects, kerac2024effects}. In educational contexts, the demand for more personalized and interactive 3D avatars is growing rapidly, as learners seek deeper immersion and more expressive modes of engagement. However, despite this growing interest, current systems often fall short in providing avatars that are both realistically adaptive and pedagogically integrated. Most existing solutions \bluecite{falloon2010using, soliman2013implementing} rely on static, pre-defined models with limited capacity for personalization, emotional nuance, or real-time responsiveness. This gap highlights the need for advanced frameworks that support expressive, high-fidelity avatar interactions, aligned with individual learner identities and educational goals. Fortunately, recent advancements in deep generative models have made this increasingly feasible. Modern avatar-generation pipelines, powered by Latent Diffusion Models (LDMs) \bluecite{cao2024dreamavatar}, Generative Adversarial Networks (GANs) \bluecite{koh2024systematic}, and Neural Radiance Fields (NeRFs) \bluecite{zhao2023havatar}, enable the synthesis of avatars from diverse inputs such as text, images, and geometric descriptors. Moreover, student engagement remains a cornerstone of effective learning, particularly within these evolving online and immersive environments. Research consistently shows that higher levels of student engagement, encompassing behavioral, emotional, and cognitive dimensions, strongly correlate with improved learning outcomes, enhanced knowledge retention, and increased course completion rates \bluecite{dixson2015measuring}. To promote and maintain such engagement, learning analytics has emerged as an essential tool, providing educators and platforms with detailed insights into learner behaviors and progress \bluecite{ifenthaler2019utilising, ouyang2024ai}. By analyzing interaction data, learning analytics systems can identify disengagement early and enable timely, personalized interventions that support learner motivation and success. Although platforms like Open edX and Moodle have incorporated analytics frameworks, the integration between learning analytics and advanced generative AI-driven personalization is still limited. Addressing this integration gap is crucial for creating adaptive educational experiences that dynamically respond to individual learner engagement patterns, thus maximizing the potential of immersive and AI-powered learning environments. Together, these advances underscore the importance of developing integrated, flexible educational platforms that use AI and immersive technologies to foster personalized, engaging, and effective learning experiences.

To address these intersecting challenges, we present Open TutorAI, an open-source educational platform designed to provide learners with a personalized and interactive support powered by LLMs and generative technologies. The platform combines advanced natural language processing models to power interactive learning experiences, including AI-generated tutors and customizable 3D avatars. It features tools for course organization, integrated feedback systems, and distinct interfaces for students, educators, and parents. The platform includes a structured onboarding process to identify each learner’s preferences, goals, and support needs. Based on this input, a personalized AI assistant is dynamically configured to support the learner with explanations, practice activities, and contextual guidance. Learners can engage with the assistant through a traditional chat interface or a more immersive 3D avatar experience, enhancing the sense of presence and interactivity. Unlike static e-learning systems, Open TutorAI offers students control over their help, encouraging autonomy, motivation, and adaptability with different learning styles. 

The key contributions of this work are:
\begin{itemize}
  \item An open-source, LLM-powered educational platform that offers adaptive and interactive tutoring experiences, including both chat-based and avatar-based modalities.
  \item A customizable assistant-generation workflow that configures support based on each learner’s goals, preferences, and needs.
  \item A user-friendly interface that encourages learner autonomy and personalization without requiring technical expertise.
  \item Integration of generative avatar capabilities to support immersion and identity expression in virtual learning environments.
  \item Learning analytics features for tracking student progress and engagement over time.
\end{itemize}

The remainder of this paper is organized as follows: Section \ref{sec2} reviews related work on conversational agents, avatar-based interfaces, large language model integration, and learning analytics in educational contexts. Section \ref{sec3} outlines the proposed solution, detailing the system architecture, user interface, core functionalities, and extensibility features of Open TutorAI. Section \ref{sec4} presents the technical validation of the platform, including proof of concept, experimental design, and deployment considerations. Section \ref{sec5} examines both the pedagogical and technological implications of the proposed system and presents future perspectives and directions for further development. Finally, Section \ref{sec6} concludes the paper.

\section{Related work}
\label{sec2}
\subsection{Conversational Agents and Educational Chatbots}
The intersection of artificial intelligence and education has led to the development of numerous tools aimed at enhancing the learning experience through personalization, adaptability, and automation \bluecite{kurni2023beginner}. Early contributions in this space include Intelligent Tutoring Systems (ITS), which provide individual support by modeling student knowledge and delivering tailored feedback \bluecite{jimenez2018affective}. These systems laid the foundation for adaptive learning platforms, such as Smart Sparrow \bluecite{polly2014evaluation} and Desire2Learn (D2L), both of which adapt content delivery based on learner interactions. A prominent example is Jill Watson, a virtual teaching assistant developed using IBM Watson, which was deployed in online computer science courses to respond to student queries and reduce instructor load \bluecite{goel2018jill}. Despite their innovative approach, such assistants often lacked the flexibility to adapt to individual learner needs and were confined to specific courses or institutions.

Building on these developments, our earlier work explored the integration of advanced NLP techniques to improve educational chatbots. We utilized Xatkit, a chatbot development framework, and proposed an encoder-decoder architecture for intent recognition, employing CamemBERT to encode utterances and an intent classification decoder to detect student intents \bluecite{tarek2022towards}. In subsequent work, we implemented and tested a chatbot in public schools to support the teaching of an educational programming language \bluecite{ait2024impact}. Results showed improved learner motivation, reduced stress, and greater autonomy. Despite these advances, such systems were still limited by rigid conversational flows and predefined intents.

LLMs have subsequently transformed this field by allowing for more dynamic, context-sensitive, and personalized learning experiences that surpass static responses \bluecite{maity2024generative}. Tools like Khanmigo by Khan Academy use GPT-powered agents to simulate Socratic-style tutoring, helping students with problem-solving in real time \bluecite{shetye2024evaluation}. Similarly, commercial applications such as Duolingo Max integrate LLMs for conversational language learning and feedback. Table \ref{tab:RtabAll} highlights key differences between Open TutorAI and existing commercial LLM-powered educational platforms. While these platforms demonstrate the educational potential of generative AI, most are closed-source, lack transparency, and offer limited customization for individual learners or institutions.

\begin{table*}[htb]
\centering
\footnotesize
\caption{Comparison of AI-powered educational platforms}
\renewcommand{\arraystretch}{1.5} 
\begin{tabular}{>{\centering\arraybackslash}m{3cm}
                >{\centering\arraybackslash}m{2cm}
                >{\centering\arraybackslash}m{5cm}
                >{\centering\arraybackslash}m{3cm}
                >{\centering\arraybackslash}m{3cm}} 
\hline\hline
\textbf{Platform} & \textbf{Open Source} & \textbf{Personalization} & \textbf{Multimodality} & \textbf{Integrated Analytics} \\ \hline\hline
Khanmigo (Khan Academy) & Closed & Medium & Text only & Limited \\ 
Duolingo Max & Closed & Medium & Text + Audio & Limited \\ 
Open TutorAI & Open  & High (user-configurable paths), Avatar customization, Adaptive tutoring & Text + 3D avatar & Yes (learner dashboards \& engagement analytics) \\ \hline\hline
\end{tabular}
\label{tab:RtabAll}
\end{table*}

\subsection{Interfaces for Large Language Model Integration in Educational Platforms
}

The development of extensible interfaces for managing LLMs has become a central element in modern educational technologies, enabling adaptive, multi-model interactions tailored to diverse learner needs. Such interfaces support personalized dialogue, real-time content generation, and seamless integration with pedagogical frameworks. One of the most versatile solutions in this domain is OpenWebUI \bluecite{baek_designing_2024}, an open-source, self-hosted platform originally developed as a front-end for the Ollama LLM runner. It has since evolved into a flexible orchestration layer capable of managing both API-based and local LLMs, including OpenAI’s GPT-4, as well as open-source alternatives like Qwen and ERNIE.

OpenWebUI’s modular architecture supports dynamic model switching, parallel querying, and ensemble response merging, making it particularly well-suited for experimentation and personalization. Its compatibility with containerized environments (e.g., Docker, Kubernetes) ensures accessibility and ease of deployment, even for users without advanced technical expertise. Beyond text interaction, the platform offers multimodal capabilities such as voice and video interfaces, vision-enabled model support, and integration with image-generation frameworks, features that align closely with the needs of immersive and adaptive learning environments.

In the context of Open TutorAI, OpenWebUI serves as the core orchestration layer, facilitating the management of dynamic tutoring dialogues, AI behavior personalization, and secure communication with multiple LLM backends. Its extensibility enables the integration of structured prompt pipelines, avatar-based interfaces, and RAG modules. This foundation supports the delivery of inclusive, learner-centered educational experiences that are both scalable and adaptable across domains and languages.

\subsection{Learning Analytics and Behavioral Engagement in Intelligent Tutoring Systems}

Student engagement is consistently recognized as a critical factor influencing academic success, particularly in the online learning environment \bluecite{hu2017student}. It encompasses behavioral, emotional, and cognitive dimensions, all of which contribute to the depth of learner interaction, persistence, and achievement. Recent studies confirm that higher engagement levels are associated with improved learning outcomes, increased course completion rates, and greater learner satisfaction \bluecite{sailer2020gamification}. As educational technologies evolve, the integration of socially interactive elements, such as real-time feedback, gamified learning experiences, embodied avatars, and personalized video-based recommendations has been shown to significantly boost cognitive and emotional engagement \bluecite{el2023video, hennessey2025effects}. These multimodal features not only stimulate motivation but also foster a sense of presence and connection, which are essential for sustained participation and deeper learning.

To complement engagement-focused design, learning analytics has emerged as a powerful framework for monitoring, interpreting, and optimizing learning processes based on data-driven insights. Modern learning analytics systems can detect behavioral trends, identify early indicators of disengagement, and offer personalized interventions to assist learner success by evaluating user interaction data \bluecite{johnston2024uncovering}. Platforms such as Open edX and Moodle have integrated foundational analytics modules, enabling instructors to visualize learning trajectories and adapt teaching strategies accordingly \bluecite{ruiperez2017evaluation}. However, many of these systems operate independently of AI-driven pedagogical components, limiting their ability to provide actionable feedback in real time. Bridging this gap requires the seamless integration of learning analytics with generative AI technologies, enabling dynamic adaptation of learning content, tutoring strategies, and feedback based on continuously evolving engagement metrics.

\subsection{The Role of Avatars in Pedagogical Interfaces}

While LLM-powered educational tools have unlocked new levels of personalization and adaptability, an equally important dimension of learner engagement comes from the integration of embodied, multimodal interactions, particularly through the use of animated avatars. The integration of 3D animated avatars into educational tools has added an essential multimodal layer to learner interaction, enhancing engagement, emotional resonance, and social connection. Animated avatars serve not only as visual companions but also as social agents, facilitating communication, reducing learner anxiety, and improving the perceived social presence of virtual tutors \bluecite{kyrlitsias2022social}. Prior research has demonstrated that embodied conversational agents (ECAs) with lifelike gestures, facial expressions, and gaze behaviors can significantly influence learner motivation, trust, and knowledge retention \bluecite{waltemate2018impact}. For example, our prior work uses animated agents to deliver adaptive feedback, scaffold learning, and simulate human-like teaching behaviors \bluecite{tarek2022towards}. More recent work has extended this approach into virtual reality (VR) contexts, where LLM-driven non-playable characters (NPCs) engage learners using synchronized speech, gesture recognition, and gaze tracking \bluecite{el2025architecture}.  

Building on the integration of 3D animated avatars, embodied conversational agents, and LLM-powered tools, Open TutorAI provides an open-source, learner-centered platform that unites multimodal interaction with real-time personalization. The integration of animated agents with the conversational capacities of LLM in Open TutorAI enhances learner engagement and strengthens the sense of social presence within educational interactions. As an open-source platform, Open TutorAI prioritizes accessibility, transparency, and adaptability, offering learners and educators the ability to customize support according to individual needs and institutional settings, unlike many proprietary solutions that limit flexibility and openness.

\section{Solution Outline}
\label{sec3}
The design of Open TutorAI follows a research-driven approach to the development of next-generation educational tools \bluecite{eriksson2025design}. It prioritizes simplicity, scalability, and pedagogical effectiveness, ensuring that the system remains both accessible and adaptable across diverse learning contexts. Open TutorAI features a modular, open-source architecture that enables the delivery of personalized educational support by using recent advances in LLMs and RAG techniques. The system builds upon the extensible OpenWebUI framework \bluecite{baek_designing_2024}, offering a flexible foundation for integrating adaptive dialogue, dynamic content retrieval, and multimodal user interfaces.

\subsection{Architecture overview}
\begin{figure*}[htb]
    \centering
    \includegraphics[width=\textwidth]{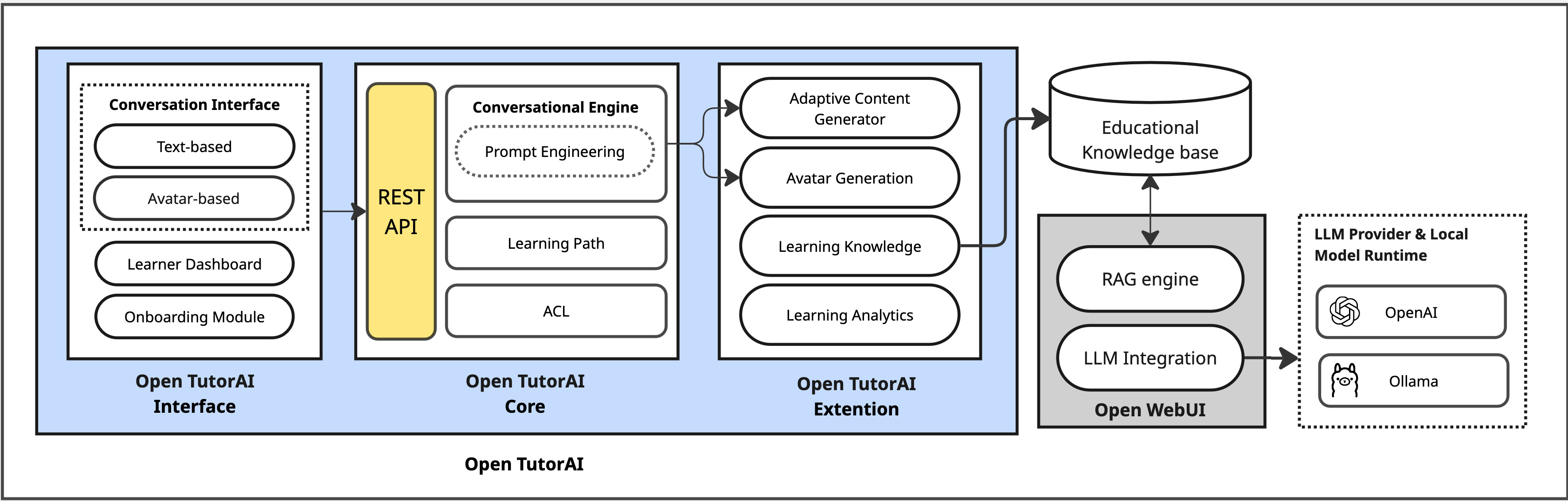}
    \caption{System architecture of Open TutorAI.}
    \label{fig:overview}
\end{figure*}

The architecture of Open TutorAI is designed to be modular, extensible, and community-friendly, promoting transparency, reusability, and ease of customization. As illustrated in Figure \ref{fig:overview}, the system is organized into three main components: the frontend interface, the core logic and extension modules, and the backend infrastructure, which includes external integrations with language models and knowledge bases.

\subsection{Open TutorAI Interface}
\begin{figure}[htb]
  \centering
  \begin{minipage}{0.45\textwidth}
    \centering
    \includegraphics[width=\textwidth]{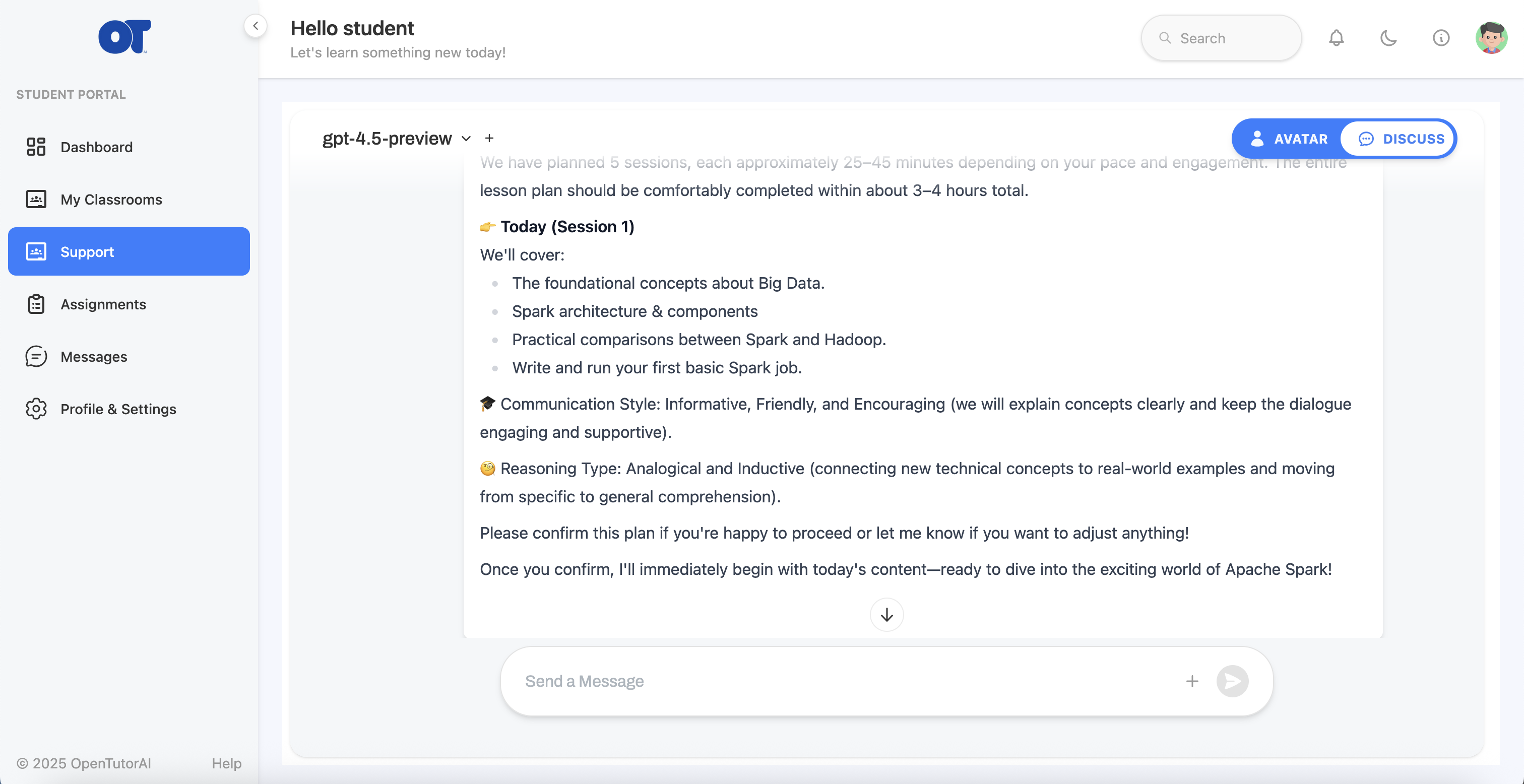}
    \vspace{3pt}
    \small (a) Text-based chat interface
  \end{minipage}
  \hfill
  \begin{minipage}{0.45\textwidth}
    \centering
    \includegraphics[width=\textwidth]{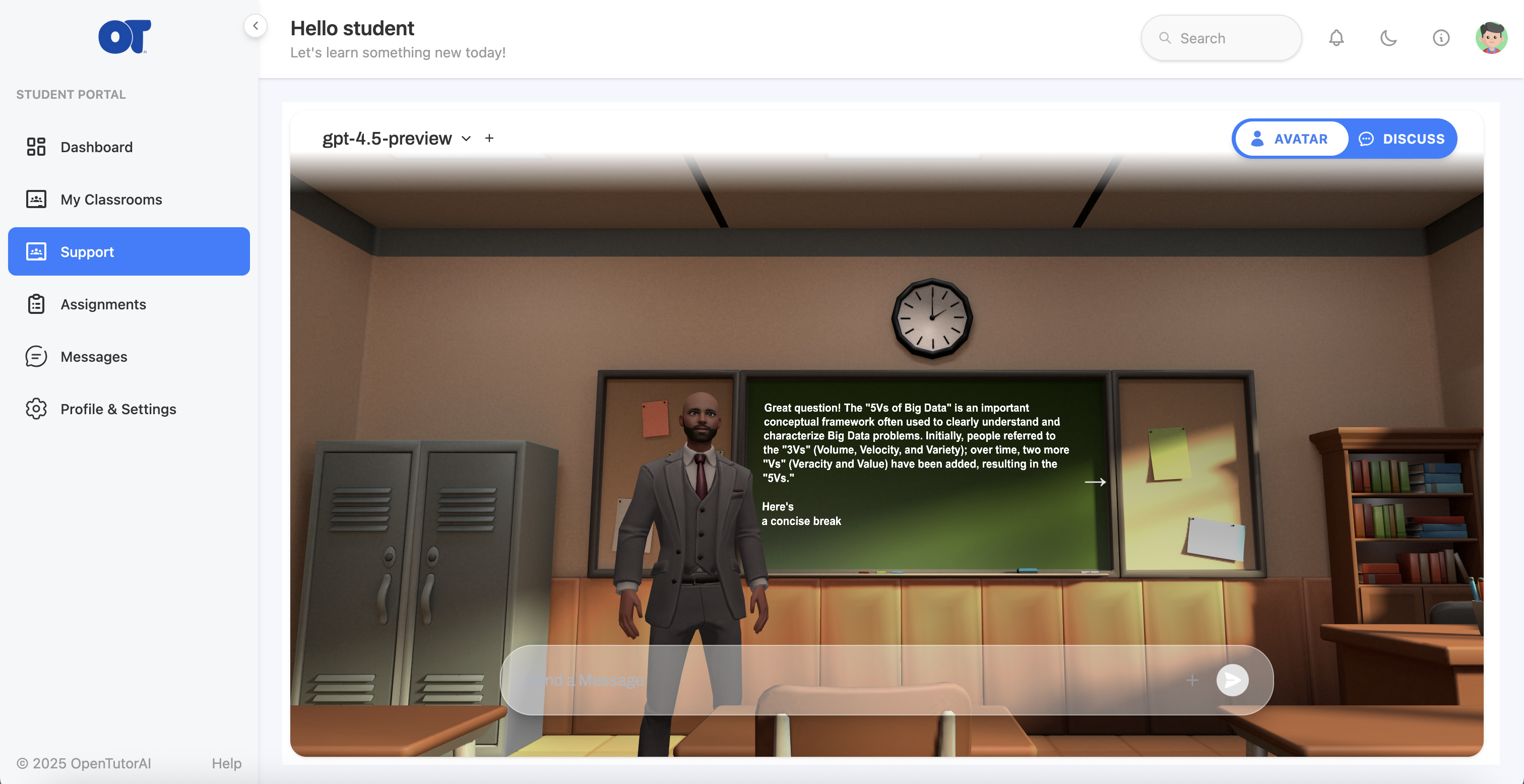}
    \vspace{3pt}
    \small (b) Avatar-based interface
  \end{minipage}
  \caption{User interfaces of Open TutorAI: (a) Text-based chat interface for direct messaging with the tutor; (b) Avatar-based 3D interface with embodied conversational agents in immersive learning environments.}
  \label{fig:combined}
\end{figure}

The Open TutorAI Interface facilitates the user experience and interaction. It includes a conversation interface supporting both text-based and avatar-based modalities, a learner dashboard for progress tracking and feedback, and an onboarding module that initializes the learner profile and preferences. Figure \ref{fig:combined} presents the user interfaces of Open TutorAI, including a text-based chat interface and an avatar-based 3D environment that supports embodied interactions. The multimodal interaction design aims to accommodate different learning styles and enhance engagement.

\subsection{Open TutorAI Core}
At the heart of the platform, the Open TutorAI Core serves as the central orchestration layer responsible for managing conversational dynamics and adaptive learning logic. This component governs the dialogue engine, leveraging prompt engineering techniques and access control mechanisms to ensure secure, context-aware interactions. It also configures personalized learning trajectories by integrating individual learner profiles with pedagogical rules and adaptation strategies. 

\begin{figure*}[htb]
    \centering
    \includegraphics[width=\textwidth]{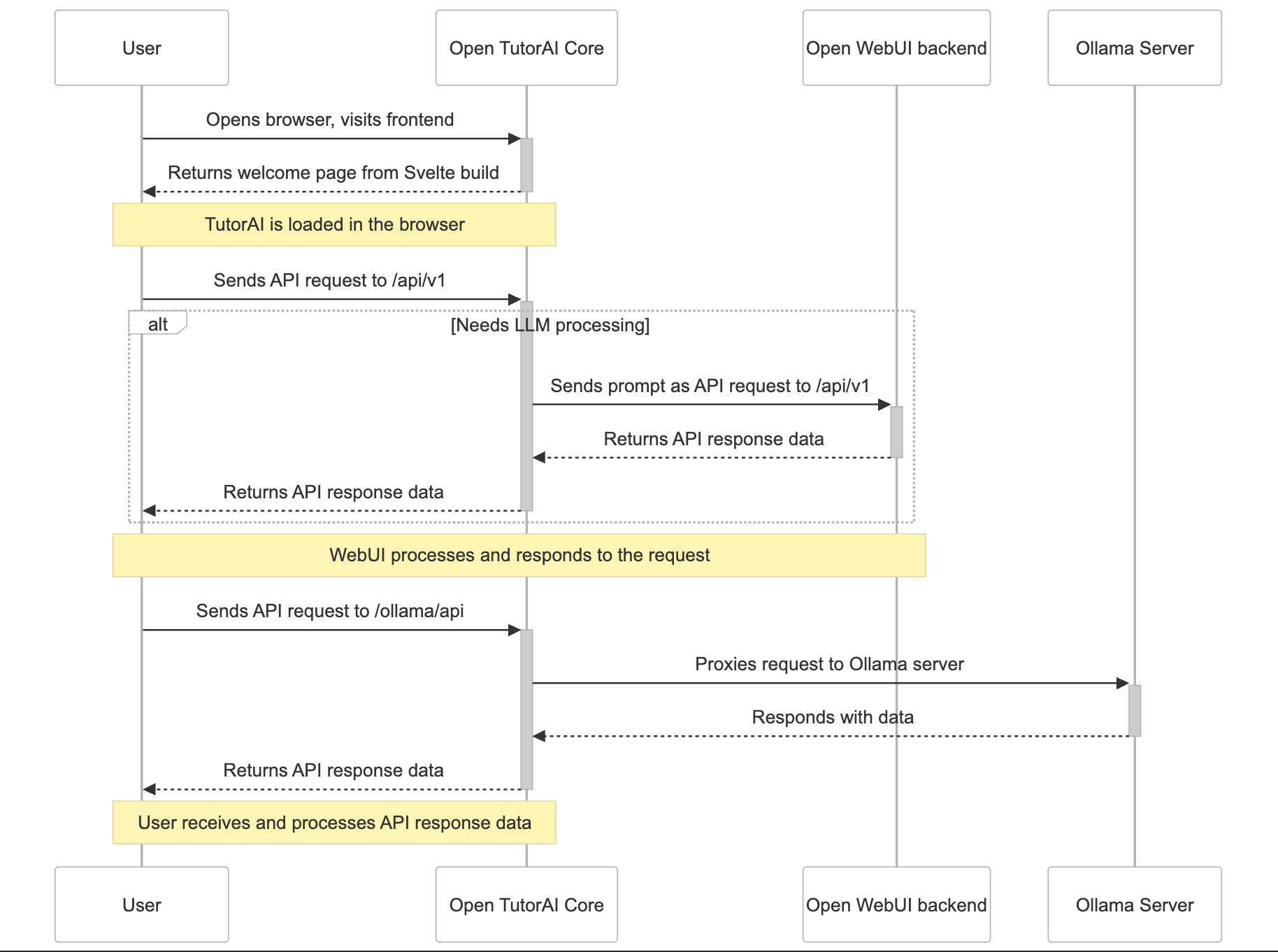}
    \caption{The macro sequence diagram of Open TutorAI, illustrating the communication flow between the user, frontend, backend (WebUI), and LLM providers (Ollama).}
    \label{fig:Sequence}
\end{figure*}

Built upon the modular infrastructure of the OpenWebUI framework, the core is designed with extensibility and interoperability in mind. This allows for seamless integration of third-party tools and facilitates continuous evolution in response to emerging technological innovations and instructional methodologies. As illustrated in Figure \ref{fig:Sequence}, the core orchestrates the communication flow to ensure that the content delivered is pedagogically aligned, contextually relevant, and dynamically adapted to the learner’s evolving needs.

The Core Component Layer encompasses the essential modules that form the foundation of the system’s functionality. It integrates the Learning Path, the Conversational Engine, and the Access Control and Role Management modules, each addressing a critical aspect of the platform’s operation. The following sections describe these components in detail.

\subsubsection{Learning Path}

\begin{figure}[htb]
    \centering
    \includegraphics[width=0.6\columnwidth]{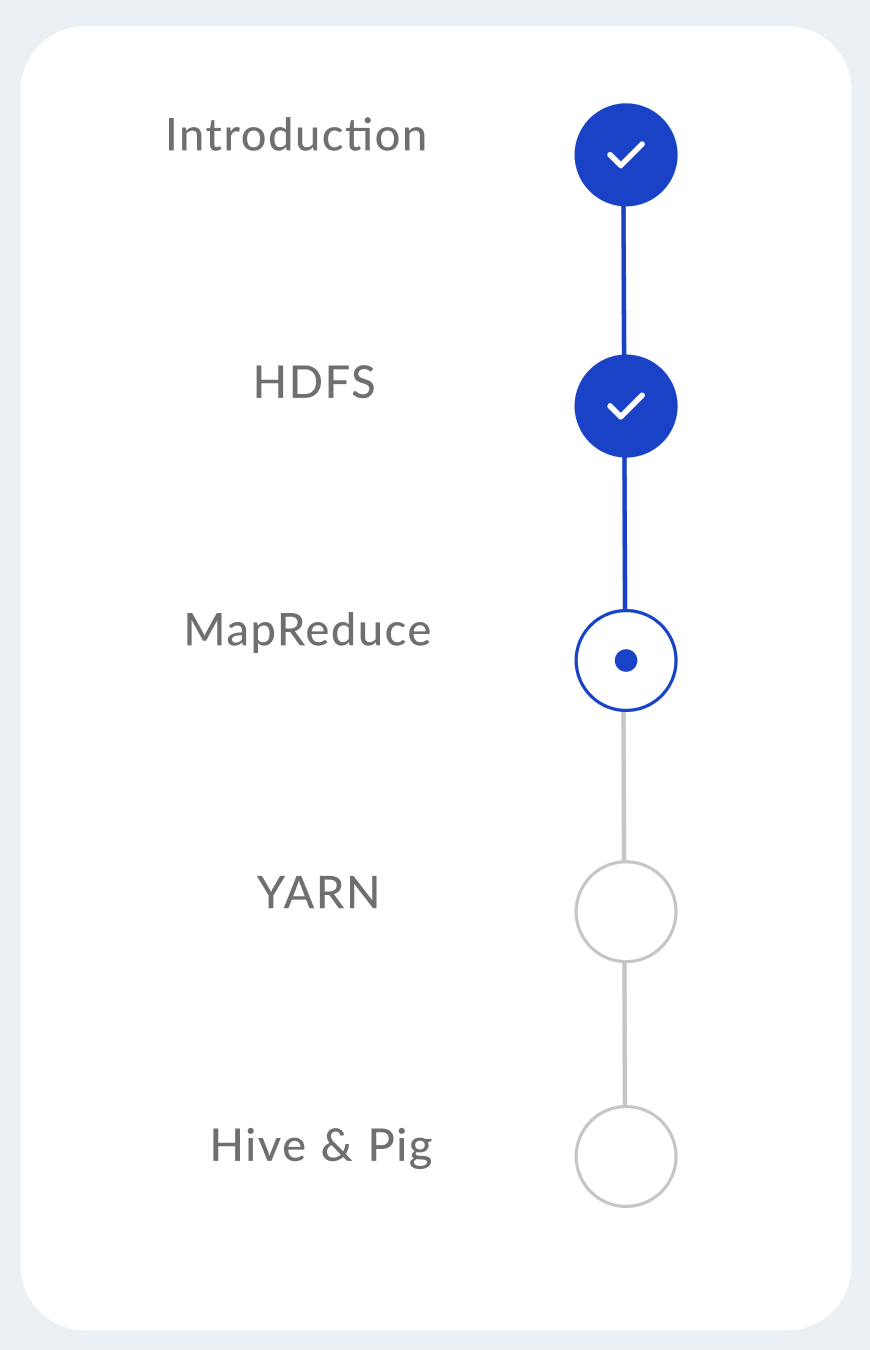}
    \caption{Learning path visualization supporting learner progress monitoring.}
    \label{fig:LP}
\end{figure}

To facilitate learner autonomy and improve navigability within the instructional process, Open TutorAI incorporates a structured learning path visualization interface, as shown in Figure \ref{fig:LP}. This component provides a graphical representation of the learner’s progression across defined pedagogical modules, enabling users to monitor their advancement, revisit prior content, and anticipate upcoming topics. For instance, in the context of distributed systems instruction, the path may span sequential modules such as Introduction, HDFS, MapReduce, and YARN, with visual cues distinguishing completed units from those in progress or pending. Such visualization serves not only to enhance user orientation but also to support self-regulated learning strategies.

\subsubsection{Conversational engine}

At the heart of Open TutorAI’s adaptive communication system is its conversational engine, which facilitates real-time, pedagogically informed interactions between the learner and the AI tutor. Rather than functioning as a simple response generator, the engine operates as an intelligent instructional orchestrator, dynamically shaping dialogue based on contextual, cognitive, and pedagogical cues. Built upon a modular architecture, it integrates learner modeling, instructional design principles, and real-time adaptation strategies to tailor each conversational exchange to the learner’s evolving profile and educational needs.

A critical component within this engine is the prompt engineering module, which acts as the adaptive intelligence layer responsible for dynamically generating, customizing, and optimizing prompts sent to the underlying LLMs. Unlike static query systems, this module is designed to contextualize, personalize, and pedagogically align every interaction based on learner profiles, historical interactions, instructional objectives, and real-time feedback.

\begin{itemize}
  \item \textbf{Generates structured lesson plans:} Upon receiving a learner’s request, the system automatically decomposes the target topic into a coherent sequence of sessions, each centered on a distinct concept. It estimates the total number of sessions required, defines the scope for the current session, and explicitly communicates this plan to the learner for approval before proceeding.
  \item \textbf{Adapts communication style dynamically:} The system selects the most appropriate communication tone (encouraging, friendly, informative, humorous, or neutral) based on the learner’s profile, the subject matter, and the learning context (casual exploration vs. exam preparation). This ensures not just informational accuracy, but also emotional resonance and engagement.
  \item \textbf{Applies context-sensitive reasoning strategies:} Depending on the subject domain, learner level, and learning objectives, the module configures the underlying reasoning framework to match cognitive demands: deductive for logic and math, inductive for empirical subjects, analogical to connect new and known ideas, causal to explain processes, and abductive for hypothesis generation and creative problem-solving. This ensures each session is cognitively aligned and effective.
  \item \textbf{Designs micro lesson flow:} For each ~25-minute session, the module structures learning around an engaging introduction, progressive buildup from basic to advanced concepts, integration of real-world examples and analogies, practical guidance for applying knowledge, recommendations for visual or multimedia aids, and a concise yet comprehensive delivery scaled to learner level, concluding with an active practice task that requires learner input before feedback. This scaffolding ensures clarity, engagement, and actionable learning outcomes.
  \item \textbf{Manages adaptive feedback and quizzing:} After each session, the module offers learners the option to proceed to the next lesson or take a focused 5-question quiz blending current and prior material. Quiz interactions are adaptive, presenting one question at a time with immediate, constructive feedback, offering optional content review for incorrect answers, and dynamically deciding whether to advance or reinforce based on learner performance.
  \item \textbf{Incorporates built-in adaptive strategies:} The module continuously monitors learner traces to adjust its approach by adapting the communication style, for example, using “Encouraging” tones during challenging moments, or by modifying reasoning methods, for example, adding analogies when concepts become abstract.
\end{itemize}

To operationalize this adaptive behavior, the system relies on a structured prompt design framework that encodes pedagogical logic directly within each interaction. The framework defines how lesson plans, reasoning strategies, tone selection, and assessment options are instantiated through dynamic prompt composition. Each prompt is generated from parameterized templates filled at runtime with contextual variables such as learner level, topic, performance history, and instructional objectives. This ensures consistent quality, personalization, and pedagogical coherence across sessions.

The prompt engineering strategy in OpenTutorAI follows a structured, parameter-driven methodology that ensures pedagogical alignment, personalization, and reproducibility. Rather than relying on static textual prompts, the system composes multi-layered instructions dynamically by integrating contextual data extracted from the learner profile, session history, and instructional objectives.

At its core, the prompting process follows a four-layer prompting architecture:
\begin{enumerate}
  \item Global Context Layer: Defines the pedagogical intent of the session (lesson generation, feedback, quiz, or review), specifies the model’s tutoring role, and sets communication constraints and expected output format.
  \item Instructional Logic Layer: Encodes the sequence of teaching actions—topic decomposition, tone selection, reasoning type, and learner confirmation—to align with instructional design principles.
  \item Adaptive Variable Layer: Dynamically injects contextual data such as learner level, progress, and preferred communication style into the template, allowing continuous personalization over time.
  \item Post-Interaction Layer: Manages follow-up activities, including quizzes, targeted reviews, and reinforcement prompts, and defines how to deliver constructive feedback or re-explanations when needed.
\end{enumerate}

This multi-layer prompting architecture transforms pedagogical metadata into structured instructions, ensuring that each model output remains coherent, context-aware, and educationally relevant. The system also employs compositional prompting, decomposing complex tutoring tasks into smaller chained prompts for planning, delivery, feedback, and evaluation. This design enhances control, interpretability, and the system’s ability to adapt in real time to learners’ evolving needs.

\subsubsection{Access control and Role management}

A critical component of Open TutorAI’s system architecture is its Access Control Layer (ACL), which governs user permissions, content access, and system interactions based on role-specific privileges. Given the multi-user nature of the platform, including students, teachers, parents and administrators, the ACL ensures a secure, context-aware, and pedagogically coherent experience for every profile.

Open TutorAI employs a role-based access control model, whereby permissions are dynamically assigned based on the user’s role within the learning ecosystem:

\begin{itemize}
  \item Learners have access to the tutoring interfaces (chat or avatar-based), personalized learning paths, progress tracking tools, and goal-setting modules. Their environment is designed to encourage autonomy, motivation, and adaptive engagement.
  \item Teachers are granted access to tools for monitoring student performance, analyzing learning analytics, configuring content delivery, and intervening with targeted feedback or support.
  \item Parents benefit from a dedicated dashboard that offers high-level insights into their child’s progress and engagement patterns. Their access is designed to promote constructive involvement while respecting learner privacy, particularly for minors.
  \item Administrators manage global platform settings, including model configurations (e.g., OpenAI or Ollama), onboarding flows, user management, and compliance with institutional and regulatory frameworks.
\end{itemize}

Technically, access control is enforced across both frontend and backend layers:

\begin{itemize}
  \item On the frontend, user interfaces are conditionally rendered based on role-specific permissions.
  \item On the backend, authorization tokens and middleware validations secure all sensitive operations, ensuring that only authorized users can retrieve or modify protected data.
\end{itemize}

This architecture ensures that Open TutorAI provides a safe, customizable, and ethically sound environment for all profiles, aligning educational functionality with privacy, security, and pedagogical integrity.

\subsection{Open TutorAI Extensions}

This module manages advanced functionalities, including the adaptive content generator that employs RAG techniques to deliver precise and contextually relevant responses. It also encompasses the avatar generation engine, which creates expressive, personalized 2D and 3D avatars aligned with user preferences. Additionally, a comprehensive learning analytics component monitors engagement metrics, tracks learner progress, and produces actionable insights to support continuous improvement. The extensions are designed with a layered, loosely coupled architecture that promotes scalability, maintainability, and ease of integration. This approach enables seamless incorporation of third-party tools, flexible substitution of LLM backends, and ongoing enhancement of pedagogical features informed by user feedback and technological advances.

The backend interfaces with external LLM providers such as OpenAI and Ollama, as well as knowledge bases through the RAG engine. This integration facilitates dynamic content retrieval and improves the factual accuracy of AI-generated responses. The architecture supports both cloud-hosted and local model deployments, providing institutions with deployment flexibility based on their technical and policy requirements.

\subsubsection{Avatar generation}
\begin{figure}[htb]
    \centering
    \includegraphics[width=\columnwidth]{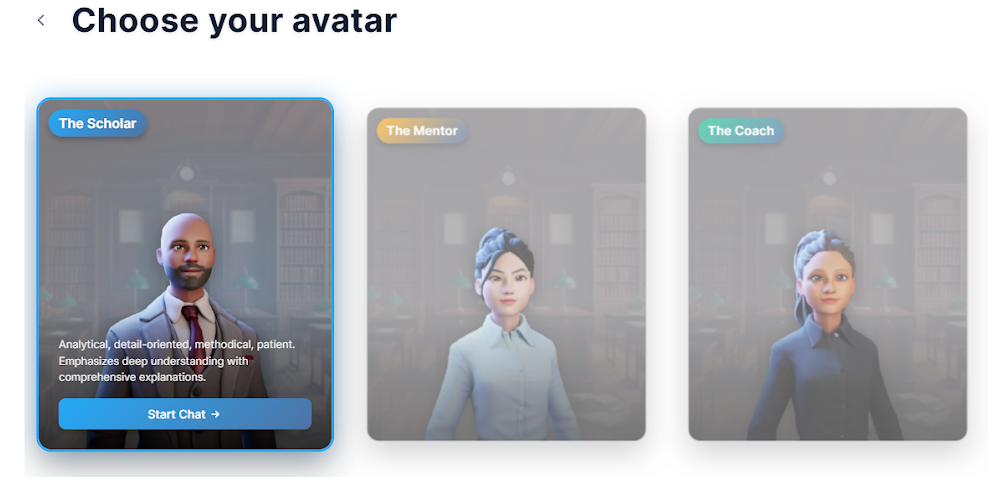}
    \caption{Avatar selection interface shown during the onboarding phase, allowing learners to personalize their virtual tutor based on preferred visual characteristics}
    \label{fig:5}
\end{figure}

The integration of a 3D animated avatar in the Open TutorAI platform is grounded in specific pedagogical objectives, aiming to address recurring challenges in digital learning environments. The incorporation of a personalized virtual tutor capable of expressing non-verbal cues enhances the communicative richness of the interaction, surpassing the limitations of text-only exchanges. As shown in Figure \ref{fig:5}, the avatar selection interface appears during the onboarding phase, allowing learners to personalize their virtual tutor according to preferred visual characteristics. This multimodal approach contributes to several key educational outcomes:

\begin{itemize}
  \item \textbf{Enhanced student engagement:} Animated avatars transform the interaction from a static, letter-like text exchange into a dynamic and responsive learning experience. This shift from passive reading to interactive communication increases deeper engagement, particularly in sustained learning sessions. 
  \item \textbf{Reinforced comprehension through non-verbal cues:} The avatar's supportive gestures, such as pointing while explaining a concept, nodding in agreement, or gesturing to highlight key points, serve to reinforce verbal explanations. These multimodal cues are particularly beneficial for visual learners, who rely on spatial and contextual signals to process information more effectively.
  \item \textbf{Enhancing personal connection and reducing anxiety:} Personalization of the tutor avatar, allowing learners to choose the avatar they want during onboarding by selecting visual traits they find appealing or comforting (e.g., friendly, trustworthy, relatable), helps foster a sense of connection and emotional safety. This personalization reduces the perceived impersonality or rigidity often associated with AI-driven systems and creates a more approachable, human-like learning interaction.
\end{itemize}

When the avatar mode is activated, learners are immersed in an interactive 3D classroom environment, which serves as the primary setting for educational presentations. The centerpiece of this environment is a large, animated 3D board where explanations from the AI tutor are visually rendered.

\begin{figure}[htb]
    \centering
    \includegraphics[width=\columnwidth]{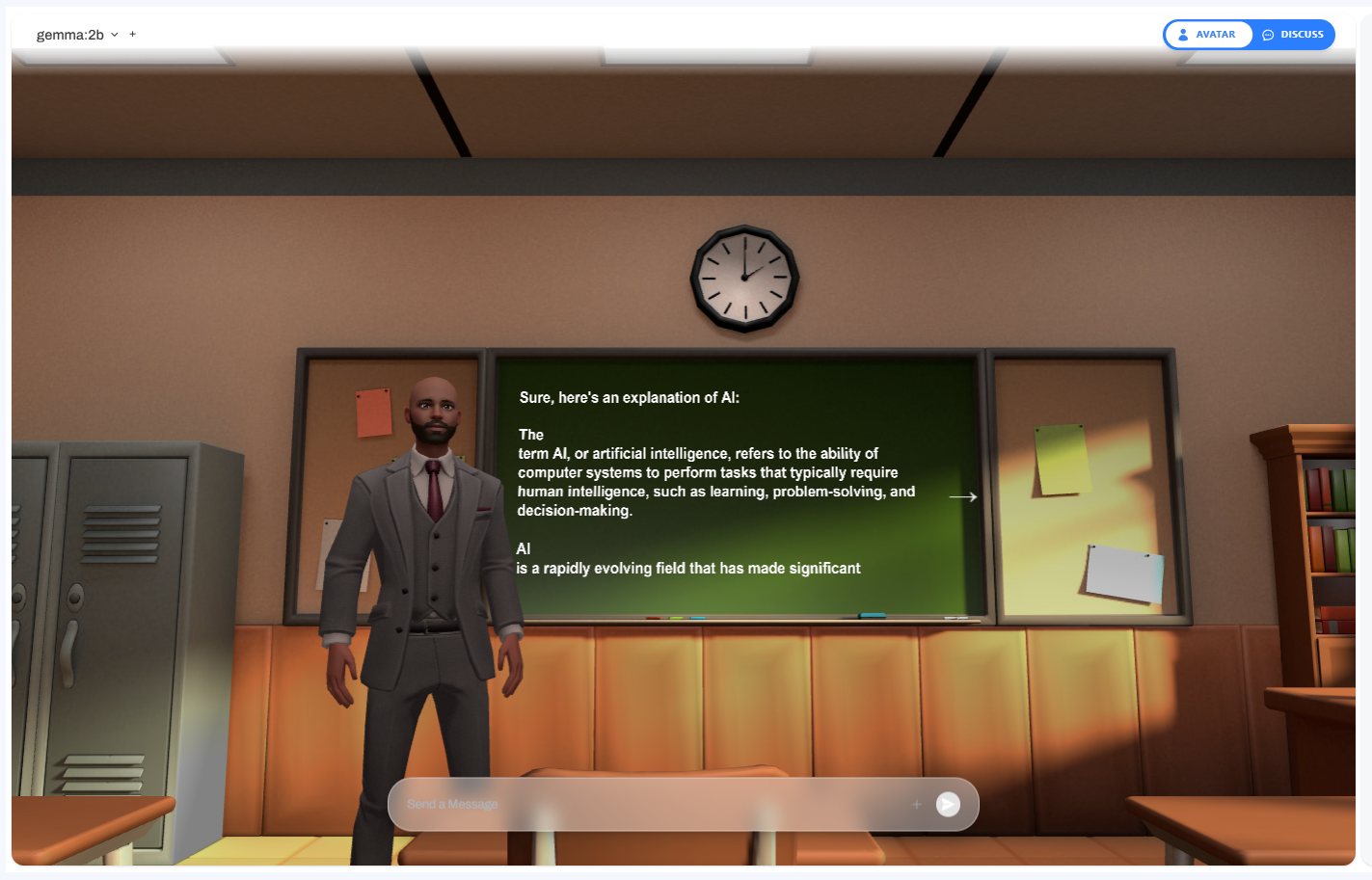}
    \caption{Avatar selection interface shown during the onboarding phase, allowing learners to personalize their virtual tutor based on preferred visual characteristics}
    \label{fig:3D}
\end{figure}

A key design challenge in developing this immersive environment was ensuring that user interface (UI) elements integrated naturally into the virtual space without disrupting the sense of presence. Instead of relying on conventional 2D overlays, the chat interface was reimagined as a floating, semi-transparent panel that appears to exist within the 3D classroom itself. This design choice preserves the spatial coherence of the environment, making interactions feel native to the virtual world rather than superimposed. The result is a unified, immersive learning experience in which learners engage with both the tutor and the interface as part of a continuous, embodied space. Figure \ref{fig:3D} illustrates this design, showing the avatar-based classroom scene with the embedded conversational UI.

To support clear content delivery in the 3D learning environment, Open TutorAI displays LLM-generated responses directly on a virtual board. An intelligent pagination system was developed to manage lengthy texts, ensuring readability by segmenting content and allowing learners to navigate through it via on-screen controls. This mechanism remains functional during text animations, preserving interactivity. Design challenges, such as maintaining text alignment during 3D board movement, were also addressed to ensure a seamless user experience.

Building on prior research that validated the pedagogical potential of immersive environments, the integration of expressive avatars and interactive 3D classrooms in Open TutorAI reflects a deliberate design choice aimed at fostering a more engaging and human-like learning experience. Earlier studies \bluecite{tarek2022towards} demonstrated that virtual agents embedded in three-dimensional educational spaces can significantly enhance learner motivation, emotional connection, and pedagogical effectiveness. By adopting these principles, Open TutorAI extends the concept of personalized tutoring into a rich, multimodal environment that aligns with contemporary advances in digital education.

\subsubsection{Learning analytics and student engagement}
The Open TutorAI platform incorporates a comprehensive learning analytics module designed to capture and analyze user interactions during the multi-step support creation process. This approach provides a detailed view of learner engagement by logging specific actions within the platform, enabling the calculation of cognitive engagement metrics based on observable behaviors. The backend system collects raw interaction data from various form fields and activities, which are then processed to derive the following key indicators:

\begin{itemize}
  \item \textbf{Learning Objective Detail \& Description Quality:} Textual content entered in the “Learning Objective” and “Short Description” fields is recorded to evaluate the learner’s clarity and depth of goal-setting, reflecting their metacognitive engagement.
  \item \textbf{Form Completion Rate \& Thoroughness:} The platform tracks the completion status of both mandatory and optional steps within the form, serving as an indicator of learner commitment and task persistence.
  \item \textbf{Time Investment per Step:} Timestamps marking when users enter and exit each step are logged to measure the duration spent, which can signify the level of cognitive focus or effort applied.
  \item \textbf{Material Upload Behavior:} Actions related to file uploads in the “Course Material” section are monitored as evidence of preparatory engagement and resource utilization.
  \item \textbf{Goal Type Preference:} User selections from the “Goal Type” options are captured to identify the nature of learner objectives, such as mastery or performance orientation.
  \item \textbf{Planning Horizon:} The chosen start and end dates for the learning objective are recorded to assess the temporal dimension of learner planning and time management.
\end{itemize}

\begin{figure*}[htb]
    \centering
    \includegraphics[width=\textwidth]{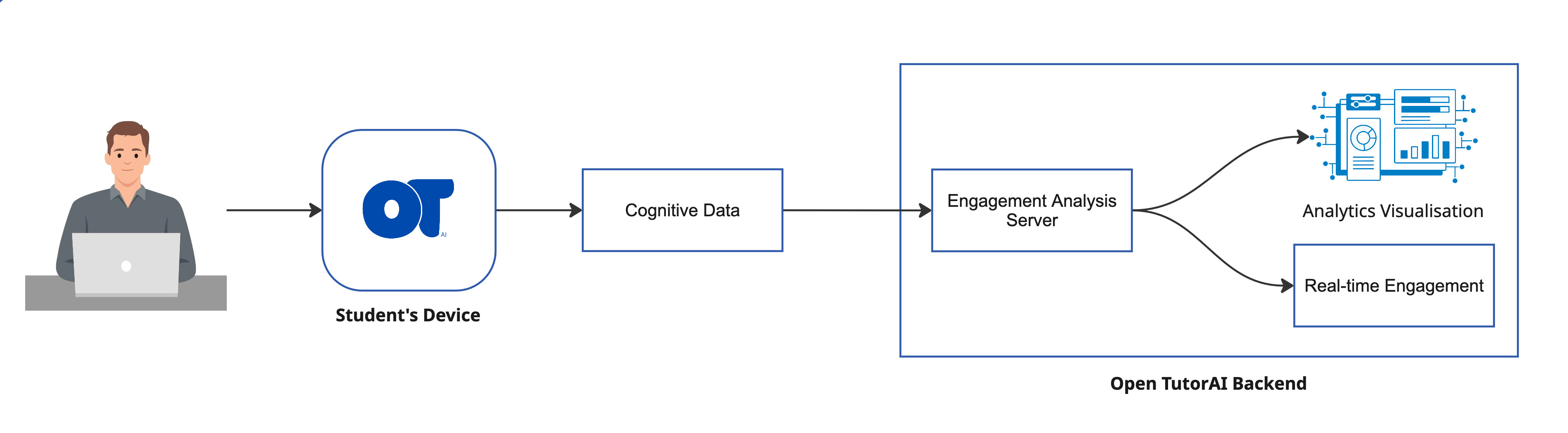}
    \caption{Data flow within Open TutorAI’s learning analytics framework: cognitive interaction data is collected from the learner’s device and transmitted to the backend engagement analysis server, where it is processed to support real-time engagement monitoring and generate visual analytics dashboards}
    \label{fig:LA}
\end{figure*}

\begin{figure*}[htb]
    \centering
    \includegraphics[width=\textwidth]{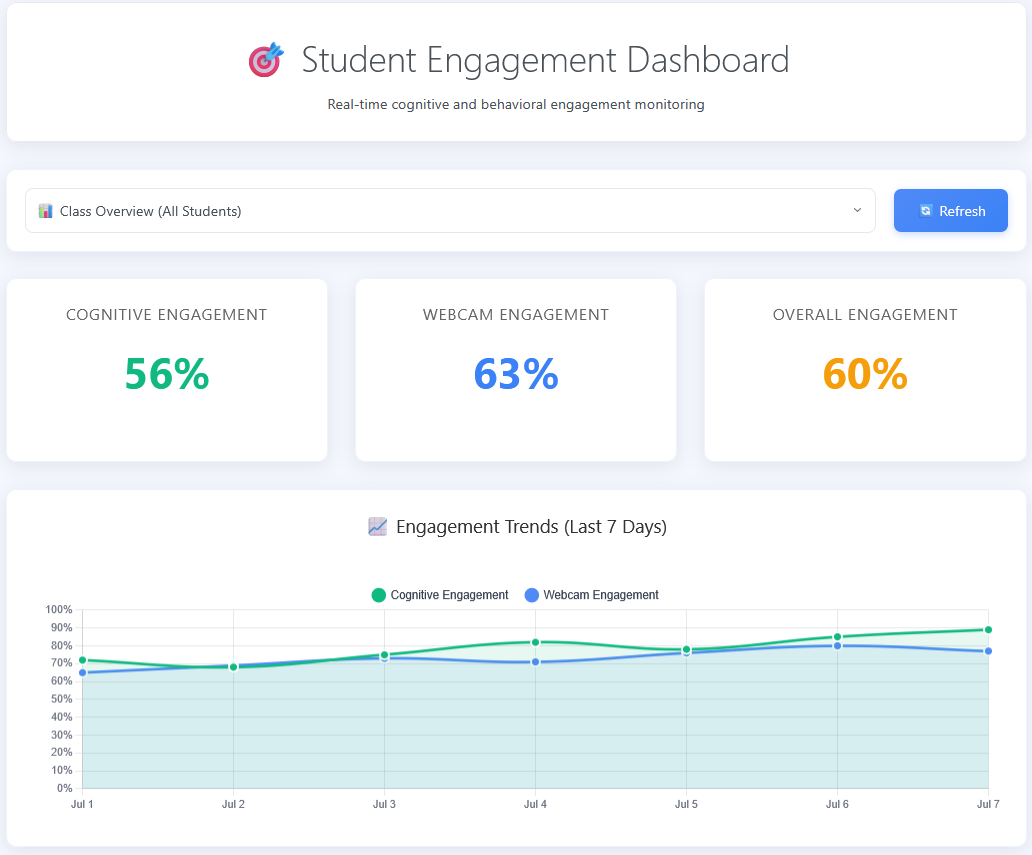}
    \caption{Real-time student engagement dashboard}
    \label{fig:Analytics}
\end{figure*}

By synthesizing these behavioral data points, the system generates a cognitive engagement score that informs adaptive interventions, supports instructor oversight, and enhances personalized tutoring strategies. This data-driven framework highlights Open TutorAI’s focus on leveraging interaction analytics to optimize learning experiences and promote sustained learner engagement. Figure \ref{fig:LA} illustrates the data flow within the engagement analysis module, from the student's device to the backend processing components responsible for real-time engagement evaluation and visual analytics. As shown in Figure \ref{fig:Analytics}, the real-time student engagement dashboard provides an overview of learners’ activity and interaction metrics throughout the session.

\subsubsection{Learning knowledge}
The Learning Knowledge module in Open TutorAI is designed to enhance educational effectiveness by integrating RAG techniques with curated, domain-specific knowledge bases. This approach ensures that AI-generated responses are not only linguistically coherent but also accurate and aligned with curricular standards. The system facilitates the delivery of content that is pedagogically based and contextually relevant by utilizing credible educational resources.

At the core of this module is a RAG engine embedded within the OpenWebUI framework, interacting with the LLM integration layer. The engine performs real-time document retrieval, assigns relevance scores, contextual fusion, and synthesizes final outputs that reflect both the learner's query and the retrieved materials. The RAG pipeline draws on a hybrid corpus that combines preloaded, educator-curated materials, such as textbooks, academic repositories, and validated open educational resources, with optional user-provided documents uploaded during learning sessions. This dual-source strategy allows the system to maintain curricular alignment while dynamically adapting to learners’ evolving needs and contexts.

This hybrid architecture enables the system to respond with high precision, making it particularly effective for domains requiring rigor, such as STEM education, academic tutoring, or professional training. Open TutorAI bridges the gap between static knowledge bases and adaptive instructional conversations by combining language generation with dynamic knowledge access. As a result, students receive better informed and trustworthy support, with current, validated insights adapted to their own learning needs.

\section{Evaluation}
\label{sec4}
The validation of Open TutorAI has been approached as a progressive process combining technical system verification, early user experience testing, and the development of a comprehensive evaluation framework. Given the novelty and modularity of the platform, spanning large language model integration, avatar-based interaction, and embedded learning analytics, each component has undergone targeted validation to ensure its educational viability and operational stability. This section presents the validation activities undertaken thus far and outlines the upcoming evaluation strategy intended to assess the platform’s pedagogical effectiveness.

\subsection{Technical Proof of Concept and system verification}

The initial validation of Open TutorAI focused on demonstrating the system’s technical reliability and readiness for deployment within realistic learning contexts. Internal testing was conducted across the full pipeline, encompassing user onboarding, learner profiling, real-time AI interaction, and behavioral engagement tracking. This testing phase aimed to ensure the coherence and stability of the system’s architecture before proceeding to broader experimental deployment.

One of the key aspects verified was the conversational engine, which successfully managed the orchestration of user prompts and adaptive responses across different language model endpoints. The system maintained responsiveness and contextual consistency, enabling fluid and personalized learning dialogues. Simultaneously, the avatar-based immersive interface was evaluated for rendering stability and interaction quality. The animated tutor avatar, integrated into a 3D classroom scene, demonstrated effective synchronization between gestures and verbal explanations, while the dynamic virtual board reliably displayed AI-generated content in a visually accessible manner. The RAG module was tested using curated educational materials. The module was able to retrieve relevant content and integrate it into coherent, accurate responses, reinforcing the factual grounding of the AI tutor’s output.

Together, these technical validations confirmed that Open TutorAI operates with scalability, and modular flexibility, ensuring that it is well-positioned for future real-world testing and iterative refinement.

\subsection{Planned experimental evaluation}

To assess the educational impact of Open TutorAI, a formal experimental study is currently being developed. The study adopts a mixed-methods approach, combining quantitative metrics and qualitative insights to evaluate how the platform influences learner engagement, knowledge acquisition, motivation, and overall satisfaction. Specifically, the evaluation aims to compare the effectiveness of the two main interaction modes of Open TutorAI (chat-based and avatar-based) against traditional edu-chatbots, providing a benchmark for the platform’s effectiveness in fostering personalized and immersive learning experiences.

The evaluation will examine learner engagement, knowledge gain, satisfaction, motivation, and trust in the virtual tutor. These dimensions will be assessed using a combination of platform analytics, pre/post tests, surveys, and user feedback, enabling a holistic understanding of the platform’s educational impact.

\subsection{System deployment and Running costs}

Open TutorAI is designed to operate efficiently across a range of deployment environments, from lightweight local setups to cloud-based infrastructures. The system’s architecture supports modular containerization, allowing it to run on standard servers with modest specifications. During development, successful deployments were achieved using Apple Silicon development machines with +16GB of RAM, enabling smooth local execution of open-source LLMs such as LLaMA 3 using the Ollama framework.

To address varying institutional needs and budgetary limitations, Open TutorAI adopts a hybrid LLM integration strategy. Through Ollama, the platform supports the deployment of open-source language models, offering a flexible, low-cost solution that can be hosted entirely on-premises. This approach is particularly advantageous for institutions prioritizing data privacy, customization, or operating in environments with limited internet connectivity, as it ensures full ownership and control over user data and system operations.

In parallel, Open TutorAI also supports commercial-grade LLMs through API-based access to providers such as OpenAI. These models (e.g., GPT-4 or GPT-4o) offer high-quality, out-of-the-box performance and advanced language capabilities, making them ideal for high-stakes educational scenarios where reliability, fluency, and multilingual support are priorities. While commercial APIs involve ongoing token-based costs, they minimize the infrastructure burden and provide seamless integration for institutions without the technical resources for local deployment.

This dual integration strategy ensures that Open TutorAI remains highly adaptable and scalable, allowing institutions to choose deployment configurations that best align with their performance requirements, budget constraints, privacy policies, and technical capacities.

Table \ref{tab:tab2} summarizes the system resource usage of Open TutorAI under two deployment configurations on an Apple Silicon machine with 16 GB RAM. Local deployment using Ollama incurs significantly higher CPU and RAM usage due to on-device model inference, while cloud-based GPT-4 offloads computation, reducing local resource consumption. Importantly, 3D avatar animation remained consistently smooth (~120 FPS) across both configurations, confirming that rendering performance is largely independent of the LLM backend. These results highlight the flexibility of Open TutorAI, allowing institutions to balance performance, privacy, and cost according to operational needs.

\begin{table*}[htb]
\centering
\footnotesize
\caption{End-to-end evaluation of Open TutorAI}
\renewcommand{\arraystretch}{1.5} 
\begin{tabular}{>{\centering\arraybackslash}m{2cm}
                >{\centering\arraybackslash}m{2.5cm}
                >{\centering\arraybackslash}m{2.5cm}
                >{\centering\arraybackslash}m{2.5cm}
                >{\centering\arraybackslash}m{1.5cm}
                >{\centering\arraybackslash}p{4.5cm}} 
\hline\hline
 & \textbf{LLM Backend} & \textbf{Average Memory Usage (RAM)} & \textbf{CPU Utilization (\%)} & \textbf{3D Animation Fluidity (FPS)} & \textbf{Remarks} \\ \hline\hline
\multirow{2}{*}{Apple Silicon} 
 & LLaMA 3 (via Ollama) 
 & \textasciitilde4 - 5 GB 
 & \textasciitilde505\%  
 & \multirow{2}{*}{\textasciitilde120 FPS} 
 & {\centering
     \vspace{-\topsep}%
     \begin{itemize}\footnotesize
        \item High local inference load
        \item Full offline operation and data privacy
        \item Stable 3D animation
     \end{itemize}} \\ \cline{2-4}\cline{6-6}
 & GPT-4 (API-based) 
 & \textasciitilde84 MB 
 & \textasciitilde94\% 
 &  
 & {\centering
     \vspace{-\topsep}%
     \begin{itemize}\footnotesize
        \item Low local computation
        \item Requires internet
        \item Minimal RAM usage
        \item High-quality language output
        \item Stable 3D animation
     \end{itemize}} \\ \hline\hline
\end{tabular}
\label{tab:tab2}
\end{table*}

\section{Discussion}
\label{sec5}
The development and implementation of Open TutorAI demonstrate how open-source, AI-powered platforms can bridge the gap between personalized learning experiences and scalable educational technologies. Through its modular architecture and hybrid integration of LLMs, Open TutorAI brings together adaptability, accessibility, and pedagogical value in a unified learning environment.

One of the platform’s most notable contributions is its ability to support multimodal interactions. While conventional chatbots in education primarily rely on scripted, text-based exchanges, Open TutorAI offers both text and 3D avatar-based modalities. This dual-mode interaction framework responds to diverse learner preferences and engagement styles, offering a more inclusive and immersive learning experience. The avatar-based classroom, for instance, creates a sense of presence that may reduce learner anxiety and increase emotional connection, key factors for long-term engagement and retention.

According to recent studies, the integration of RAG has been shown to enhance the factual reliability of responses by grounding AI-generated content in domain-specific knowledge sources \bluecite{bechard2024reducing, kumar2025improving}. These findings suggest that RAG-based systems effectively reduce factual inconsistencies and mitigate hallucinations compared to LLM-only configurations. Consequently, the system is expected to provide guidance that is both trustworthy and aligned with curricular standards. In parallel, the system’s built-in learning analytics infrastructure provides real-time behavioral insights, enabling the identification of disengagement patterns and the possibility for adaptive interventions. These capabilities collectively highlight the platform’s potential as a responsive and data-informed educational assistant.

Open TutorAI addresses key challenges related to infrastructure flexibility and system scalability by adopting a hybrid deployment strategy. The platform supports the integration of both open-source language models, through local deployment via Ollama and commercial APIs such as those provided by OpenAI. This architectural flexibility allows educational institutions to configure the system based on available resources, privacy requirements, and infrastructural constraints. Such adaptability is particularly advantageous for settings with limited internet connectivity or constrained technical capacity, facilitating broader adoption across diverse educational contexts, including under-resourced regions.

From a pedagogical perspective, Open TutorAI promotes learner autonomy by enabling users to configure their learning pathways and select their preferred mode of interaction with the AI assistant. This high degree of personalization supports increased learner motivation and engagement, while also aligning with established educational frameworks that prioritize learner-centered design, self-regulated learning, and differentiated instruction. The platform offers a more effective and inclusive learning environment by taking into account each user's choices and needs.

While initial validation confirms the operational soundness and technical readiness of the system, planned experimental evaluations will be crucial in assessing the real-world impact of Open TutorAI on learning outcomes, motivation, and satisfaction. These evaluations will also help determine the added value of immersive features, such as the 3D avatar classroom, compared to conventional interfaces.

In summary, Open TutorAI constitutes a significant contribution to the advancement of AI-enhanced educational technologies. Its design, which integrates flexibility, personalized support, and open infrastructure, provides a robust foundation for future research and large-scale deployment across varied educational settings. Ongoing development efforts, coupled with systematic empirical evaluations, will be essential for further enhancing the platform’s capabilities and establishing its effectiveness as a next-generation intelligent tutoring system.

Building upon these findings and the system’s current capabilities, it is important to look ahead toward the next stages of Open TutorAI’s evolution. While the present work establishes a solid foundation for scalable, adaptive, and data-informed tutoring, the long-term vision extends beyond technical feasibility toward a more holistic educational ecosystem. The following perspective outlines the planned directions for future development, focusing on enhancing personalization, multimodality, pedagogical alignment, and immersive learning to further advance the platform’s potential as an intelligent and human-centered tutoring system.

The current version lays the foundation for a personalized, intelligent educational platform. However, our long-term vision involves expanding its capabilities across several dimensions, including personalization, multimodal interaction, pedagogical alignment, and immersive learning. The following roadmap outlines the key development milestones foreseen in future iterations of the system.

The next iteration of the platform will focus on improving personalization and refining the overall user experience. A first step in this direction is the customization of the tutor avatar, allowing users to adjust attributes such as the name, voice, and tone. This aims to foster greater familiarity and emotional connection between the learner and the system. In parallel, enhancements to the user interface (UI) and user experience (UX) are planned. These include the addition of contextual tooltips, optimization of the layout, and simplification of navigation pathways to reduce cognitive load and increase accessibility. Furthermore, an engagement analytics dashboard will be introduced to monitor learner interactions, including click paths, feedback instances, and drop-off points. These insights will serve both to inform system refinement and to support educators in understanding learner behavior. To facilitate interoperability, we also intend to define and expose a set of application programming interfaces (APIs), allowing seamless integration with external learning management systems and third-party educational tools.

Beyond UI personalization, we aim to extend the interaction paradigm to include multimodal capabilities and multi-agent tutoring. This phase will focus on synchronizing input and feedback from various communication channels (voice, webcam, and textual input) enabling more natural and human-like exchanges \bluecite{ouarka2024deep}. A central objective is the definition of tutoring agent roles and interaction logic, whereby different virtual agents may assume complementary functions (e.g., explainer, motivator, evaluator) within a coordinated tutoring scenario. The implementation of this framework will lay the groundwork for richer pedagogical strategies and adaptive collaboration between learner and system.

As the system develops, we envisage a plugin-based design to enhance extensibility. This approach will empower educators and developers to integrate new features or domain-specific content without changing the core platform. A well-defined adapter interface will ensure compatibility across components. In parallel, while the system already supports interaction with multiple LLMs, the next phase of development will center on refining the prompt engineering pipeline. This component plays a critical role in shaping the quality, tone, and pedagogical relevance of LLM-generated responses. This refinement will further enhance the adaptability, interpretability, and instructional effectiveness of the tutoring system.

A significant pedagogical enhancement will involve aligning the tutoring system with existing curriculum. This includes the construction of a knowledge graph mapping learning objectives to educational content and user progression pathways \bluecite{abu2024systematic}. To further improve the system’s contextual responsiveness and instructional precision, we plan to incorporate RAG techniques. This will further support adaptive scaffolding, whereby the system adjusts its instructional strategies based on real-time assessments of learner understanding and progress.

As part of our medium to long-term vision, we plan to extend the tutoring system into immersive learning environments using virtual and augmented reality (VR/AR). Unlike the current system, which already incorporates a 3D animated avatar for interaction, this next phase will support immersive spatial worlds where learners can navigate, interact, and collaborate in real-time. Our development efforts will focus on enabling intuitive multimodal interaction through gesture recognition, gaze tracking \bluecite{mikhailenko2022eye}, and voice control, enhancing the sense of presence and agency within the learning environment \bluecite{el2025architecture}. To ensure the pedagogical value of this immersive approach, we will conduct comprehensive user studies to assess its impact on engagement, cognitive immersion, and learning outcomes.
    
Alongside these developments, we also aim to improve the platform's learner feedback and visual analytics \bluecite{susnjak2022learning}. Building upon the foundational analytics features of the current system, future iterations will consolidate diverse data streams (quiz performance, interaction histories, and engagement behaviors) into dynamic, user-friendly dashboards. These tools will serve both learners and educators. Learners will be able to track their own progress and control their learning styles, while educators will gain insights that will help them plan interventions and provide personalized education. To complement this data-driven approach, we plan to incorporate gamified elements designed to maintain motivation and increase engagement. The introduction of features such as achievement badges, progress milestones, and streak-based rewards will serve to reinforce consistent learning habits and celebrate effort over time.

Looking further ahead, a pivotal axis of our long-term vision involves the evolution of the platform from a reactive conversational system toward an agentic AI paradigm \bluecite{kamalov2504evolution}. In this future-oriented model, the AI tutor transitions from a passive respondent to an autonomous pedagogical agent capable of initiating, planning, and managing educational interventions proactively. Rather than waiting for learner queries, the system will be able to identify gaps, anticipate learning needs, and autonomously suggest resources, activities, or pedagogical strategies tailored to the learner's trajectory \bluecite{zha2025mentigo}. This agentic behavior will be guided by continuous monitoring of learner progress, emotional state, and cognitive engagement, enabling the tutor to act with intentionality and contextual awareness. However, such autonomy also raises crucial questions regarding supervision, pedagogical alignment, and ethical responsibility. To address these issues, the integration of Human-in-the-Loop (HITL) mechanisms will remain essential \bluecite{kumar2024applications}, ensuring that educators and domain experts maintain oversight over the AI’s decisions and interventions. This hybrid model positions the human as a guiding authority, validating, orienting, and refining the agent’s actions, thus reinforcing our long-term vision of an intelligent tutoring system that is not only adaptive and proactive but also trustworthy, aligned, and pedagogically sound.

\section{Conclusion}
\label{sec6}
This paper presented Open TutorAI, an open-source, modular educational platform that leverages large language models, generative AI, and immersive technologies to deliver adaptive, learner-centered support. Grounded in pedagogical research and designed for flexibility and accessibility, the system introduces a multimodal learning experience through chat-based interactions and personalized 3D avatars. Open TutorAI tackles major drawbacks of conventional edu-chatbot systems, including low engagement, lack of personalization, and real-time feedback, by fusing conversational AI with retrieval-augmented generation and learning analytics.

Through the use of context-aware support, emotionally supportive, visually stimulating learning environments, and the ability to choose their preferred interaction modes, the platform gives learners the ability to actively direct their educational path. The integration of learning analytics further enhances this adaptivity by enabling data-driven insights into learner behavior and engagement. While the system is still in active development, its architecture lays the groundwork for scalable, ethically aligned, and pedagogically effective AI-driven education. 

Finally, Open TutorAI contributes to the evolution of intelligent tutoring systems by blending emerging technologies with sound educational principles, offering a promising direction for the future of personalized digital learning.

\section*{Acknowledgements}
This work was supported by the Ministry of Higher Education, Scientific Research and Innovation, the Digital Development Agency (DDA), and the CNRST of Morocco (Al-Khawarizmi program, Project 22).

The authors also gratefully acknowledge the BDIA Master's students (Polydisciplinary Faculty of Taroudant, Ibnou Zohr University, Taroudant, Morocco) for their open-source contributions to the development of the initial MVP, in particular Abdallah Ougoummad, Reda El Bettioui, Abdellatif Laghjaj, and Hamid Elaaly.

\section*{Authors contributions}
\textbf{M.E.H.}: Conceptualization, Methodology, Software, Validation, Resources, Writing – Original Draft, Supervision, Project Administration, Funding Acquisition. \textbf{T.A.B.}: Methodology, Validation, Resources, Writing – Review, Supervision. \textbf{A.D.}: Software, Writing – Review \& Editing. \textbf{H.F.}: Review \& Editing. \textbf{Y.E.}: Methodology, Validation, Resources, Writing – Review, Supervision.

\bibliographystyle{IEEEtran} 
\bibliography{Ref}

@article{abu2024systematic,
  title={A systematic literature review of knowledge graph construction and application in education},
  author={Abu-Salih, Bilal and Alotaibi, Salihah},
  journal={Heliyon},
  volume={10},
  number={3},
  year={2024},
  publisher={Elsevier}
}

@article{tarek2022towards,
  title={Towards highly adaptive edu-chatbot},
  author={Tarek, AIT and El Hajji, Mohamed and Youssef, ES-SAADY and Fadili, Hammou},
  journal={Procedia Computer Science},
  volume={198},
  pages={397--403},
  year={2022},
  publisher={Elsevier}
}

@article{ait2024impact,
  title={The impact of educational chatbot on student learning experience},
  author={Ait Baha, Tarek and El Hajji, Mohamed and Es-Saady, Youssef and Fadili, Hammou},
  journal={Education and Information Technologies},
  volume={29},
  number={8},
  pages={10153--10176},
  year={2024},
  publisher={Springer}
}

@article{el2023video,
  title={Video-based learning recommender systems: A systematic literature review},
  author={El Aouifi, Houssam and El Hajji, Mohamed and Es-Saady, Youssef and Douzi, Hassan},
  journal={IEEE Transactions on Learning Technologies},
  volume={17},
  pages={485--497},
  year={2023},
  publisher={IEEE}
}

@article{bechard2024reducing,
  title={Reducing hallucination in structured outputs via Retrieval-Augmented Generation},
  author={B{\'e}chard, Patrice and Ayala, Orlando Marquez},
  journal={arXiv preprint arXiv:2404.08189},
  year={2024}
}

@inproceedings{cao2024dreamavatar,
  title={Dreamavatar: Text-and-shape guided 3d human avatar generation via diffusion models},
  author={Cao, Yukang and Cao, Yan-Pei and Han, Kai and Shan, Ying and Wong, Kwan-Yee K},
  booktitle={Proceedings of the IEEE/CVF conference on computer vision and pattern recognition},
  pages={958--968},
  year={2024}
}

@article{dixson2015measuring,
  title={Measuring student engagement in the online course: The Online Student Engagement scale (OSE).},
  author={Dixson, Marcia D},
  journal={Online Learning},
  volume={19},
  number={4},
  pages={n4},
  year={2015},
  publisher={ERIC}
}

@article{el2025architecture,
  title={An Architecture for Intelligent Tutoring in Virtual Reality: Integrating LLMs and Multimodal Interaction for Immersive Learning},
  author={El Hajji, Mohamed and Ait Baha, Tarek and Berka, Anas and Ait Nacer, Hassan and El Aouifi, Houssam and Es-Saady, Youssef},
  journal={Information},
  volume={16},
  number={7},
  pages={556},
  year={2025},
  publisher={MDPI}
}

@article{eriksson2025design,
  title={Design-based research in human--computer interaction: a scoping review},
  author={Eriksson, Eva and Elif Baykal, G{\"o}k{\c{c}}e and Torgersson, Olof},
  journal={Interacting with Computers},
  pages={iwaf030},
  year={2025},
  publisher={Oxford University Press}
}

@article{baek_designing_2024,
	title = {Designing an open-source {LLM} interface and social platforms for collectively driven {LLM} evaluation and auditing},
	language = {en},
	author = {Baek},
	year = {2024},
}

@article{falloon2010using,
  title={Using avatars and virtual environments in learning: What do they have to offer?},
  author={Falloon, Garry},
  journal={British Journal of Educational Technology},
  volume={41},
  number={1},
  pages={108--122},
  year={2010},
  publisher={Wiley Online Library}
}

@inproceedings{ferguson2014innovative,
  title={Innovative pedagogy at massive scale: teaching and learning in MOOCs},
  author={Ferguson, Rebecca and Sharples, Mike},
  booktitle={European conference on technology enhanced learning},
  pages={98--111},
  year={2014},
  organization={Springer}
}

@article{gabrielli2020chatbot,
  title={A chatbot-based coaching intervention for adolescents to promote life skills: pilot study},
  author={Gabrielli, Silvia and Rizzi, Silvia and Carbone, Sara and Donisi, Valeria and others},
  journal={JMIR human factors},
  volume={7},
  number={1},
  pages={e16762},
  year={2020},
  publisher={JMIR Publications Inc., Toronto, Canada}
}

@incollection{goel2018jill,
  title={Jill Watson: A virtual teaching assistant for online education},
  author={Goel, Ashok K and Polepeddi, Lalith},
  booktitle={Learning engineering for online education},
  pages={120--143},
  year={2018},
  publisher={Routledge}
}

@article{hennessey2025effects,
  title={Effects of Immersive Learning Technologies on Learning and Engagement for Military Chaplains},
  author={Hennessey, Megan J and Haglund, Erica and Ross, Audrey and Brent, Linda},
  journal={Innovative Higher Education},
  volume={50},
  number={1},
  pages={181--195},
  year={2025},
  publisher={Springer}
}

@article{hepperle2022aspects,
  title={Aspects of visual avatar appearance: self-representation, display type, and uncanny valley},
  author={Hepperle, Daniel and Purps, Christian Felix and Deuchler, Jonas and W{\"o}lfel, Matthias},
  journal={The visual computer},
  volume={38},
  number={4},
  pages={1227--1244},
  year={2022},
  publisher={Springer}
}

@inproceedings{hu2017student,
  title={Student engagement in online learning: A review},
  author={Hu, Min and Li, Hao},
  booktitle={2017 international symposium on educational technology (ISET)},
  pages={39--43},
  year={2017},
  organization={IEEE}
}

@incollection{ifenthaler2019utilising,
  title={Utilising learning analytics for study success: Reflections on current empirical findings},
  author={Ifenthaler, Dirk and Mah, Dana-Kristin and Yau, Jane Yin-Kim},
  booktitle={Utilizing learning analytics to support study success},
  pages={27--36},
  year={2019},
  publisher={Springer}
}

@article{islam2015learning,
  title={E-learning challenges faced by academics in higher education},
  author={Islam, Nurul and Beer, Martin and Slack, Frances},
  journal={Journal of Education and Training Studies},
  volume={3},
  number={5},
  pages={102--112},
  year={2015},
  publisher={Redfame}
}

@book{jimenez2018affective,
  title={Affective feedback in intelligent tutoring systems: A practical approach},
  author={Jim{\'e}nez, Samantha and Ju{\'a}rez-Ram{\'\i}rez, Reyes and Castillo, Victor H and Armenta, Juan Jos{\'e} Tapia},
  year={2018},
  publisher={Springer}
}

@article{johnston2024uncovering,
  title={Uncovering Student Engagement Patterns in Moodle with Interpretable Machine Learning},
  author={Johnston, Laura J and Griffin, Jim E and Manolopoulou, Ioanna and Jendoubi, Takoua},
  journal={arXiv preprint arXiv:2412.11826},
  year={2024}
}

@article{kamalov2504evolution,
  title={Evolution of ai in education: Agentic workflows. arXiv 2025},
  author={Kamalov, F and Calonge, DS and Smail, L and Azizov, D and Thadani, DR and Kwong, T and Atif, A},
  journal={arXiv preprint arXiv:2504.20082},
  year={2025}
}

@inproceedings{kerac2024effects,
  title={The Effects of Avatar Design on E-Learning: A Review},
  author={Kerac, Jelena and Golubovi{\'c}, Gala and Mili{\'c} Kereste{\v{s}}, Neda and Ili{\'c}, Tamara},
  booktitle={International Conference on Design and Digital Communication},
  pages={673--686},
  year={2024},
  organization={Springer}
}

@article{koh2024systematic,
  title={A systematic literature review of generative adversarial networks (GANs) in 3D avatar reconstruction from 2D images},
  author={Koh, Angela Jia Hui and Tan, Siok Yee and Nasrudin, Mohammad Faidzul},
  journal={Multimedia Tools and Applications},
  volume={83},
  number={26},
  pages={68813--68853},
  year={2024},
  publisher={Springer}
}

@article{kumar2025improving,
  title={Improving the reliability of LLMs: Combining CoT, RAG, self-consistency, and self-verification},
  author={Kumar, Adarsh and Kim, Hwiyoon and Nathani, Jawahar Sai and Roy, Neil},
  journal={arXiv preprint arXiv:2505.09031},
  year={2025}
}

@article{kumar2024applications,
  title={Applications, challenges, and future directions of human-in-the-loop learning},
  author={Kumar, Sushant and Datta, Sumit and Singh, Vishakha and Datta, Deepanwita and Singh, Sanjay Kumar and Sharma, Ritesh},
  journal={IEEE Access},
  volume={12},
  pages={75735--75760},
  year={2024},
  publisher={IEEE}
}

@article{spriggs2025personalizing,
  title={Personalizing Education through an Adaptive LMS with Integrated LLMs},
  author={Spriggs, Kyle and Lau, Meng Cheng and Passi, Kalpdrum},
  journal={arXiv preprint arXiv:2502.08655},
  year={2025}
}

@article{brown2020language,
  title={Language models are few-shot learners},
  author={Brown, Tom and Mann, Benjamin and Ryder, Nick and Subbiah, Melanie and Kaplan, Jared D and Dhariwal, Prafulla and Neelakantan, Arvind and Shyam, Pranav and Sastry, Girish and Askell, Amanda and others},
  journal={Advances in neural information processing systems},
  volume={33},
  pages={1877--1901},
  year={2020}
}

@book{kurni2023beginner,
  title={A beginner's guide to introduce artificial intelligence in teaching and learning},
  author={Kurni, Muralidhar and Mohammed, Mujeeb Shaik and others},
  year={2023},
  publisher={Springer Nature}
}

@article{kyrlitsias2022social,
  title={Social interaction with agents and avatars in immersive virtual environments: A survey},
  author={Kyrlitsias, Christos and Michael-Grigoriou, Despina},
  journal={Frontiers in Virtual Reality},
  volume={2},
  pages={786665},
  year={2022},
  publisher={Frontiers Media SA}
}

@article{maity2024generative,
  title={Generative ai and its impact on personalized intelligent tutoring systems},
  author={Maity, Subhankar and Deroy, Aniket},
  journal={arXiv preprint arXiv:2410.10650},
  year={2024}
}

@inproceedings{mikhailenko2022eye,
  title={Eye-tracking in immersive virtual reality for education: a review of the current progress and applications},
  author={Mikhailenko, Maria and Maksimenko, Nadezhda and Kurushkin, Mikhail},
  booktitle={Frontiers in Education},
  volume={7},
  pages={697032},
  year={2022},
  organization={Frontiers Media SA}
}

@article{mousavinasab2021intelligent,
  title={Intelligent tutoring systems: a systematic review of characteristics, applications, and evaluation methods},
  author={Mousavinasab, Elham and Zarifsanaiey, Nahid and R. Niakan Kalhori, Sharareh and Rakhshan, Mahnaz and Keikha, Leila and Ghazi Saeedi, Marjan},
  journal={Interactive Learning Environments},
  volume={29},
  number={1},
  pages={142--163},
  year={2021},
  publisher={Taylor \& Francis}
}

@article{ouarka2024deep,
  title={A deep multimodal fusion method for personality traits prediction},
  author={Ouarka, Ayoub and Ait Baha, Tarek and Es-Saady, Youssef and El Hajji, Mohamed},
  journal={Multimedia Tools and Applications},
  pages={1--23},
  year={2024},
  publisher={Springer}
}

@article{ouyang2024ai,
  title={AI-driven learning analytics applications and tools in computer-supported collaborative learning: A systematic review},
  author={Ouyang, Fan and Zhang, Liyin},
  journal={Educational Research Review},
  volume={44},
  pages={100616},
  year={2024},
  publisher={Elsevier}
}

@article{polly2014evaluation,
  title={Evaluation of an adaptive virtual laboratory environment using Western Blotting for diagnosis of disease},
  author={Polly, Patsie and Marcus, Nadine and Maguire, Danni and Belinson, Zack and Velan, Gary M},
  journal={BMC medical education},
  volume={14},
  number={1},
  pages={222},
  year={2014},
  publisher={Springer}
}

@article{al2021advantages,
  title={Advantages and disadvantages of using e-learning in university education: Analyzing students’ perspectives},
  author={Al Rawashdeh, Alaa Zuhir and Mohammed, Enaam Youssef and Al Arab, Asma Rebhi and Alara, Mahmoud and Al-Rawashdeh, Butheyna},
  journal={Electronic Journal of E-learning},
  volume={19},
  number={3},
  pages={107--117},
  year={2021}
}

@article{ruiperez2017evaluation,
  title={Evaluation of a learning analytics application for open edX platform},
  author={Ruip{\'e}rez-Valiente, Jos{\'e} A and Mu{\~n}oz-Merino, Pedro J and Pijeira D{\'\i}az, H{\'e}ctor J and Santofimia Ruiz, Javier and Delgado Kloos, Carlos and others},
  journal={Computer Science \& Information Systems},
  volume={14},
  number={1},
  pages={51--73},
  year={2017},
  publisher={ComSIS Consortium}
}

@article{sailer2020gamification,
  title={The gamification of learning: A meta-analysis},
  author={Sailer, Michael and Homner, Lisa},
  journal={Educational psychology review},
  volume={32},
  number={1},
  pages={77--112},
  year={2020},
  publisher={Springer}
}

@article{sandoval2018design,
  title={Design and implementation of a chatbot in online higher education settings.},
  author={Sandoval, Zoroayka V},
  journal={Issues in Information Systems},
  volume={19},
  number={4},
  year={2018}
}

@article{sharma2025role,
  title={The role of large language models in personalized learning: a systematic review of educational impact},
  author={Sharma, Sahil and Mittal, Puneet and Kumar, Mukesh and Bhardwaj, Vivek},
  journal={Discover Sustainability},
  volume={6},
  number={1},
  pages={1--24},
  year={2025},
  publisher={Springer}
}

@article{shetye2024evaluation,
  title={An evaluation of Khanmigo, a generative AI tool, as a computer-assisted language learning app},
  author={Shetye, Shamini},
  journal={Studies in Applied Linguistics and TESOL},
  volume={24},
  number={1},
  year={2024}
}

@inproceedings{soliman2013implementing,
  title={Implementing Intelligent Pedagogical Agents in virtual worlds: Tutoring natural science experiments in OpenWonderland},
  author={Soliman, Mohamed and Guetl, Christian},
  booktitle={2013 IEEE Global Engineering Education Conference (EDUCON)},
  pages={782--789},
  year={2013},
  organization={IEEE}
}

@article{susnjak2022learning,
  title={Learning analytics dashboard: a tool for providing actionable insights to learners},
  author={Susnjak, Teo and Ramaswami, Gomathy Suganya and Mathrani, Anuradha},
  journal={International Journal of Educational Technology in Higher Education},
  volume={19},
  number={1},
  pages={12},
  year={2022},
  publisher={Springer}
}

@article{waltemate2018impact,
  title={The impact of avatar personalization and immersion on virtual body ownership, presence, and emotional response},
  author={Waltemate, Thomas and Gall, Dominik and Roth, Daniel and Botsch, Mario and Latoschik, Marc Erich},
  journal={IEEE transactions on visualization and computer graphics},
  volume={24},
  number={4},
  pages={1643--1652},
  year={2018},
  publisher={IEEE}
}

@article{yeganeh2025future,
  title={The future of education: A multi-layered metaverse classroom model for immersive and inclusive learning},
  author={Yeganeh, Leyli Nouraei and Fenty, Nicole Scarlett and Chen, Yu and Simpson, Amber and Hatami, Mohsen},
  journal={Future Internet},
  volume={17},
  number={2},
  pages={63},
  year={2025},
  publisher={MDPI}
}

@inproceedings{zha2025mentigo,
  title={Mentigo: An Intelligent Agent for Mentoring Students in the Creative Problem Solving Process},
  author={Zha, Siyu and Liu, Yujia and Zheng, Chengbo and Xu, Jiaqi and Yu, Fuze and Gong, Jiangtao and Xu, Yingqing},
  booktitle={Proceedings of the 2025 CHI Conference on Human Factors in Computing Systems},
  pages={1--22},
  year={2025}
}

@article{zhao2023havatar,
  title={Havatar: High-fidelity head avatar via facial model conditioned neural radiance field},
  author={Zhao, Xiaochen and Wang, Lizhen and Sun, Jingxiang and Zhang, Hongwen and Suo, Jinli and Liu, Yebin},
  journal={ACM Transactions on Graphics},
  volume={43},
  number={1},
  pages={1--16},
  year={2023},
  publisher={ACM New York, NY, USA}
}

\appendix
\renewcommand{\thefigure}{A\arabic{figure}}
\setcounter{figure}{0}

\begin{figure*}[h!]
  \centering
  \includegraphics[width=\linewidth]{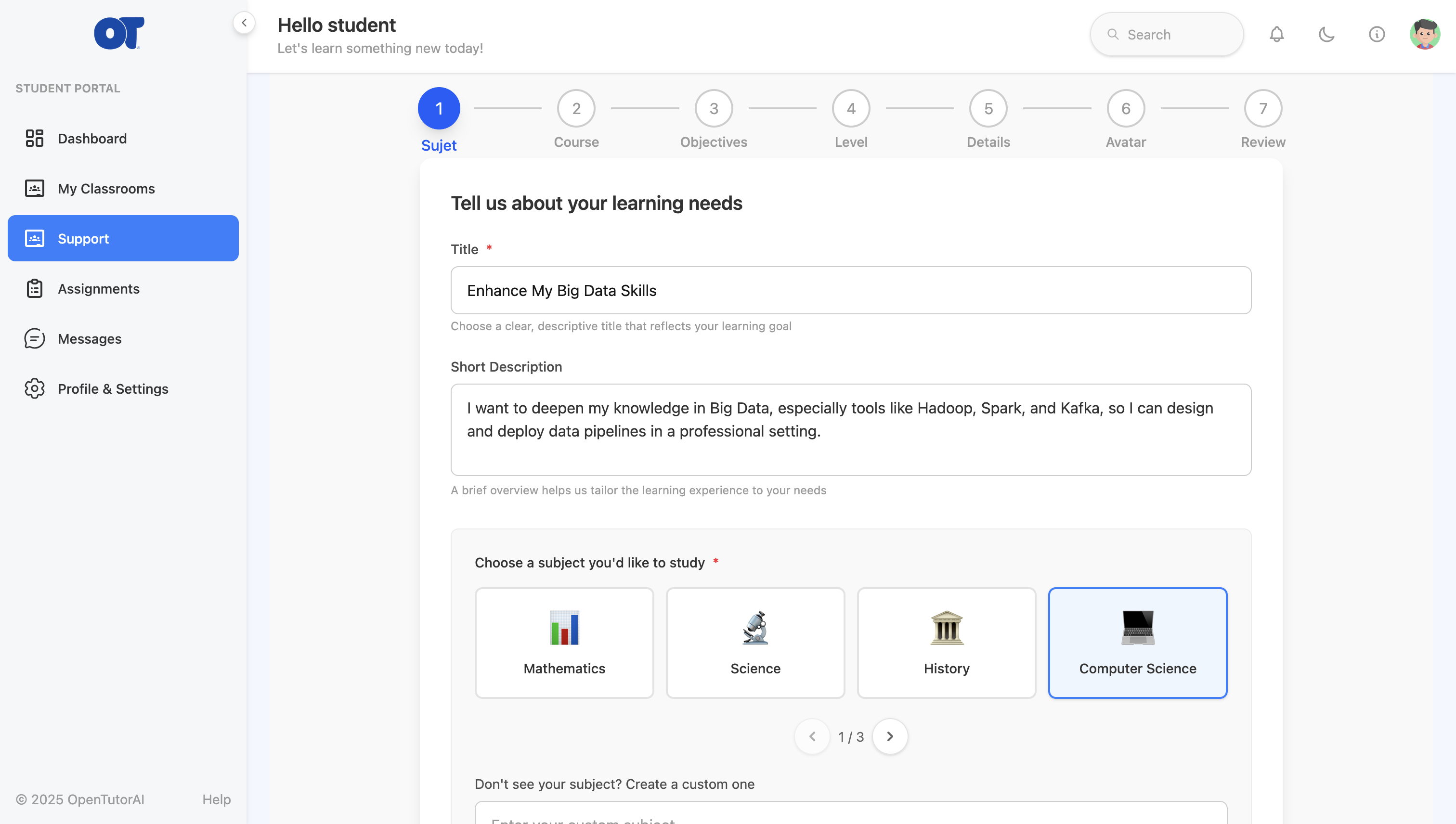}
  \caption{Step 1: The user provides a concise and specific goal and describes their motivation or context that will guide the tone and focus of the support. The user also selects a subject area (e.g. Computer Science), which helps route the request to the appropriate knowledge scope of the support. }
  \label{fig:A1}
\end{figure*}

\begin{figure*}[htb]
  \centering
  \includegraphics[width=\linewidth]{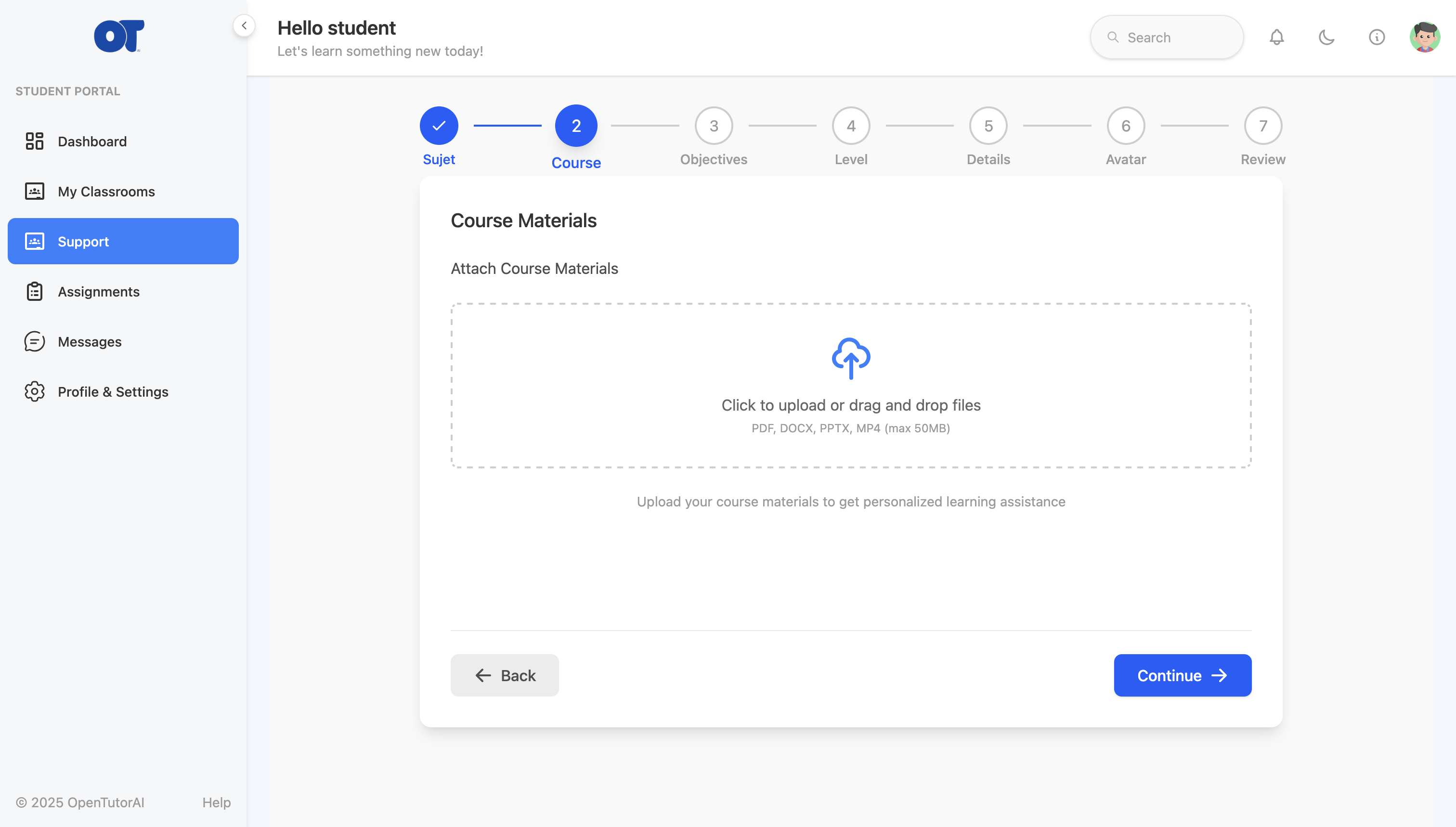}
  \caption{Step 2: The user specifies what they want to explore or achieve in this session and chooses a mode like “Build a new skill,” “Review a course,” or “Prepare for an exam”. This helps narrow the support's guidance to immediate needs and tune tone.}
  \label{fig:A2}
\end{figure*}

\begin{figure*}[htb]
  \centering
  \includegraphics[width=\linewidth]{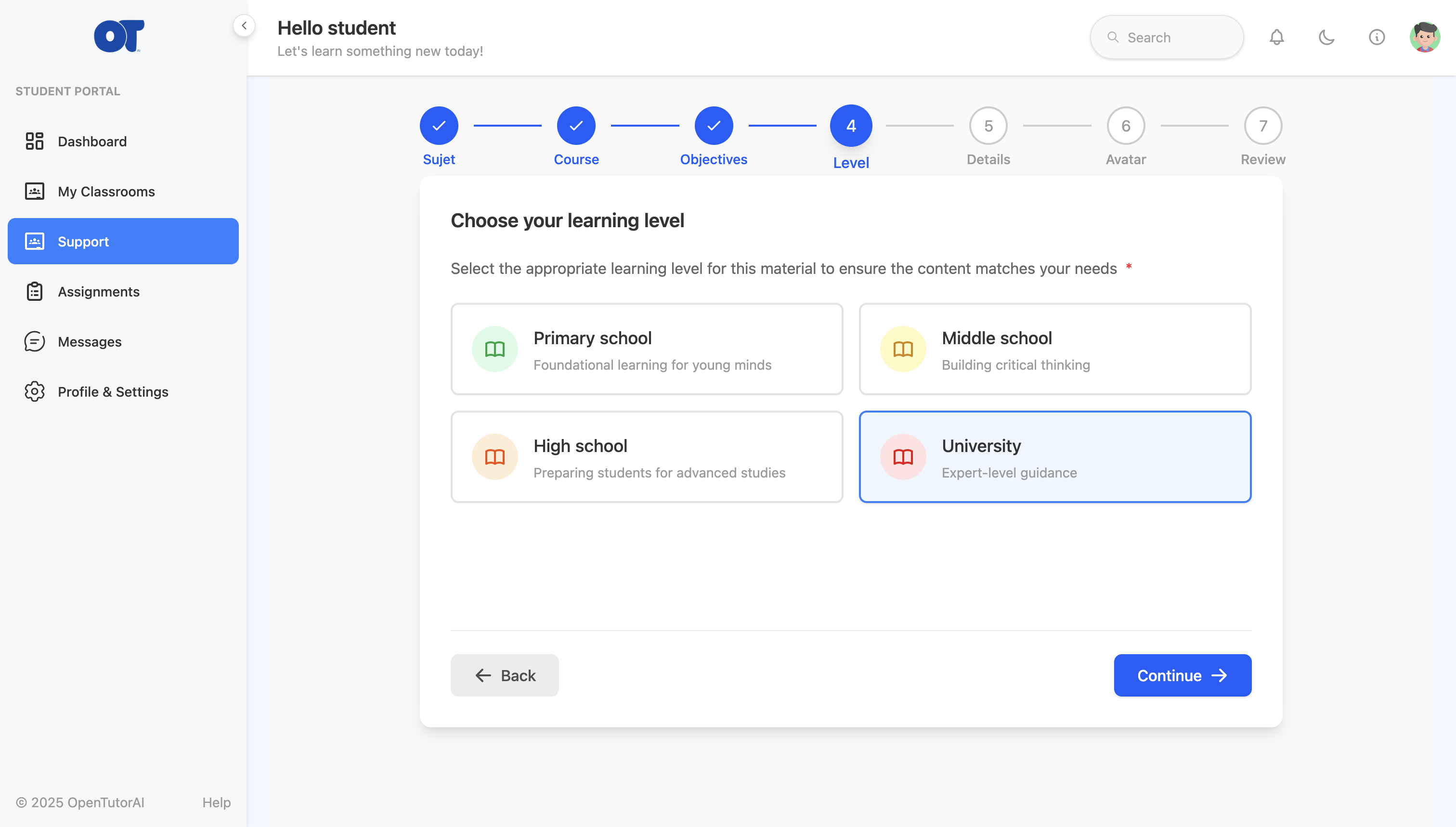}
  \caption{Step 3: The user uploads relevant course materials such as PDFs, presentations, or videos. These documents serve as the knowledge base for personalized learning support. This step ensures the support can align closely with the exact content and structure the learner is engaging with.}
  \label{fig:A3}
\end{figure*}

\begin{figure*}[htb]
  \centering
  \includegraphics[width=\linewidth]{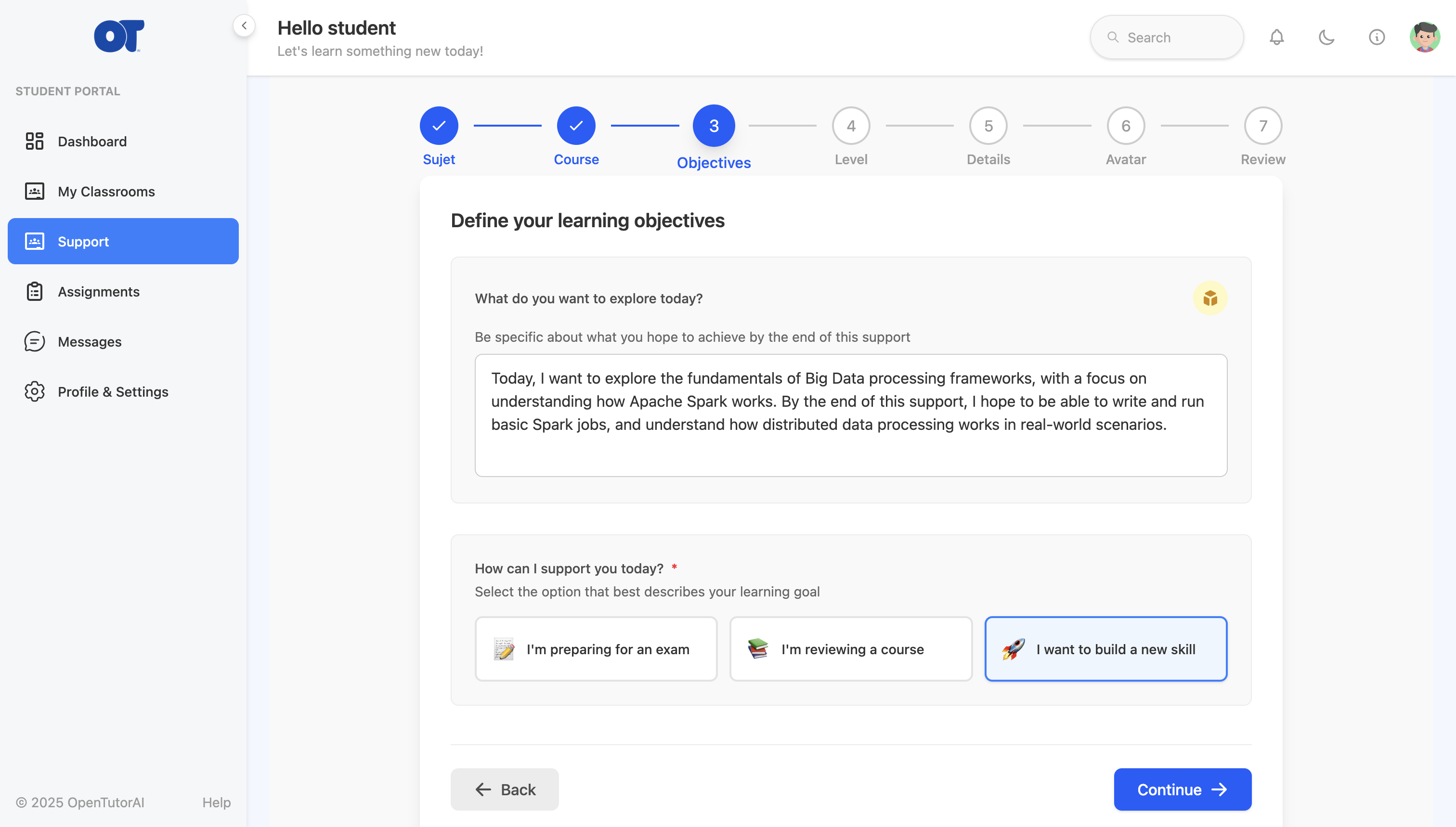}
  \caption{Step 4: The user selects their educational level from predefined categories: Primary School, Middle School, High School, or University. This classification helps the system adapt content depth, vocabulary, and approach to ensure age-appropriate guidance and effective learning.}
  \label{fig:A4}
\end{figure*}

\begin{figure*}[htb]
  \centering
  \includegraphics[width=\linewidth]{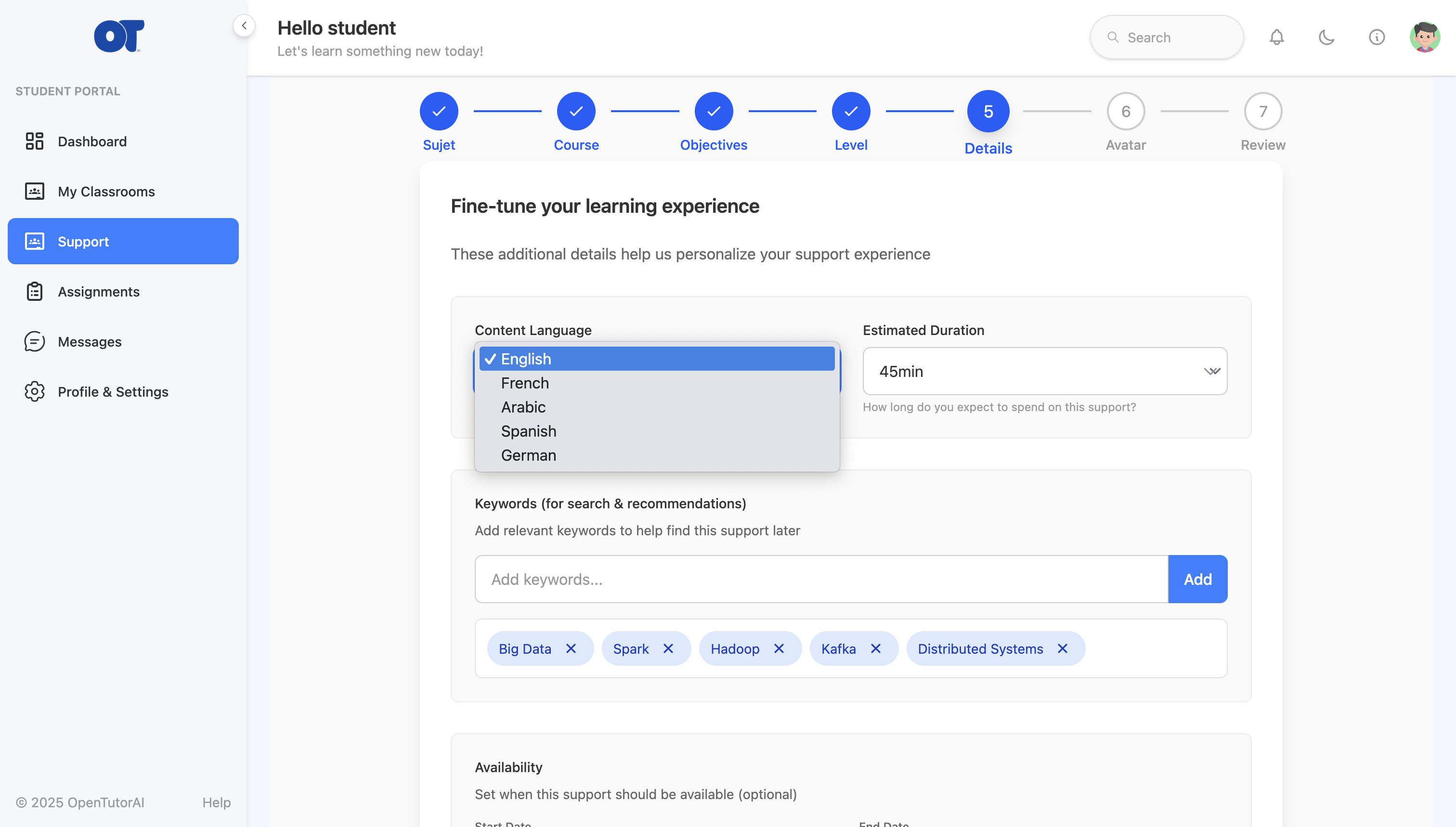}
  \caption{Step 5: The user fine-tunes their experience by selecting the content language, specifying estimated duration, adding keywords for searchability, and optionally indicating availability. This step personalizes the support experience to the user’s preferences, context, and scheduling needs.}
  \label{fig:A5}
\end{figure*}

\begin{figure*}[htb]
  \centering
  \includegraphics[width=\linewidth]{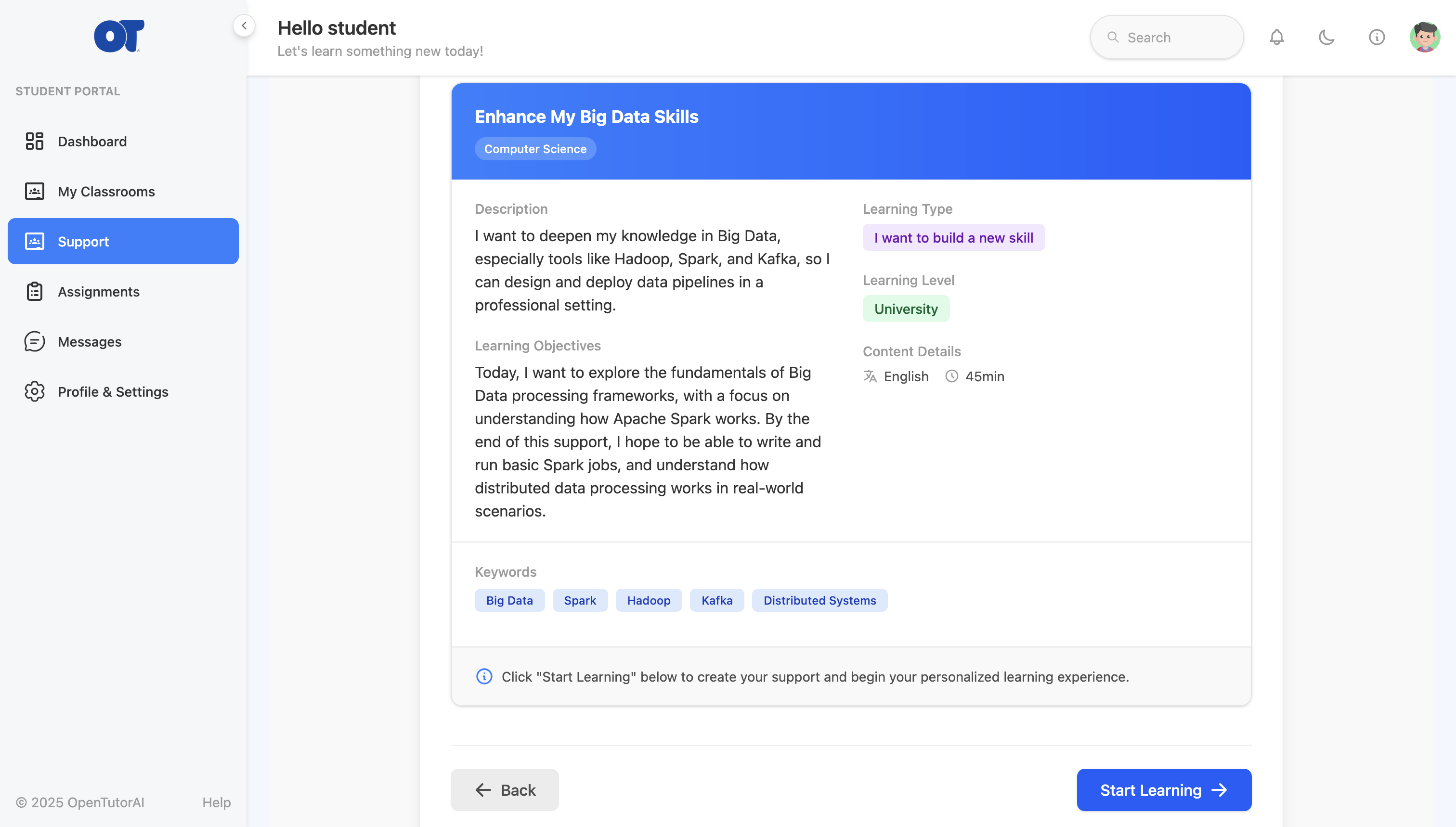}
  \caption{Step 6: The user reviews all selected inputs and customizations before submitting the support request. This ensures accuracy, alignment with expectations, and readiness for launching a tailored support session.}
  \label{fig:A6}
\end{figure*}

\begin{figure*}[htb]
  \centering
  \includegraphics[width=\linewidth]{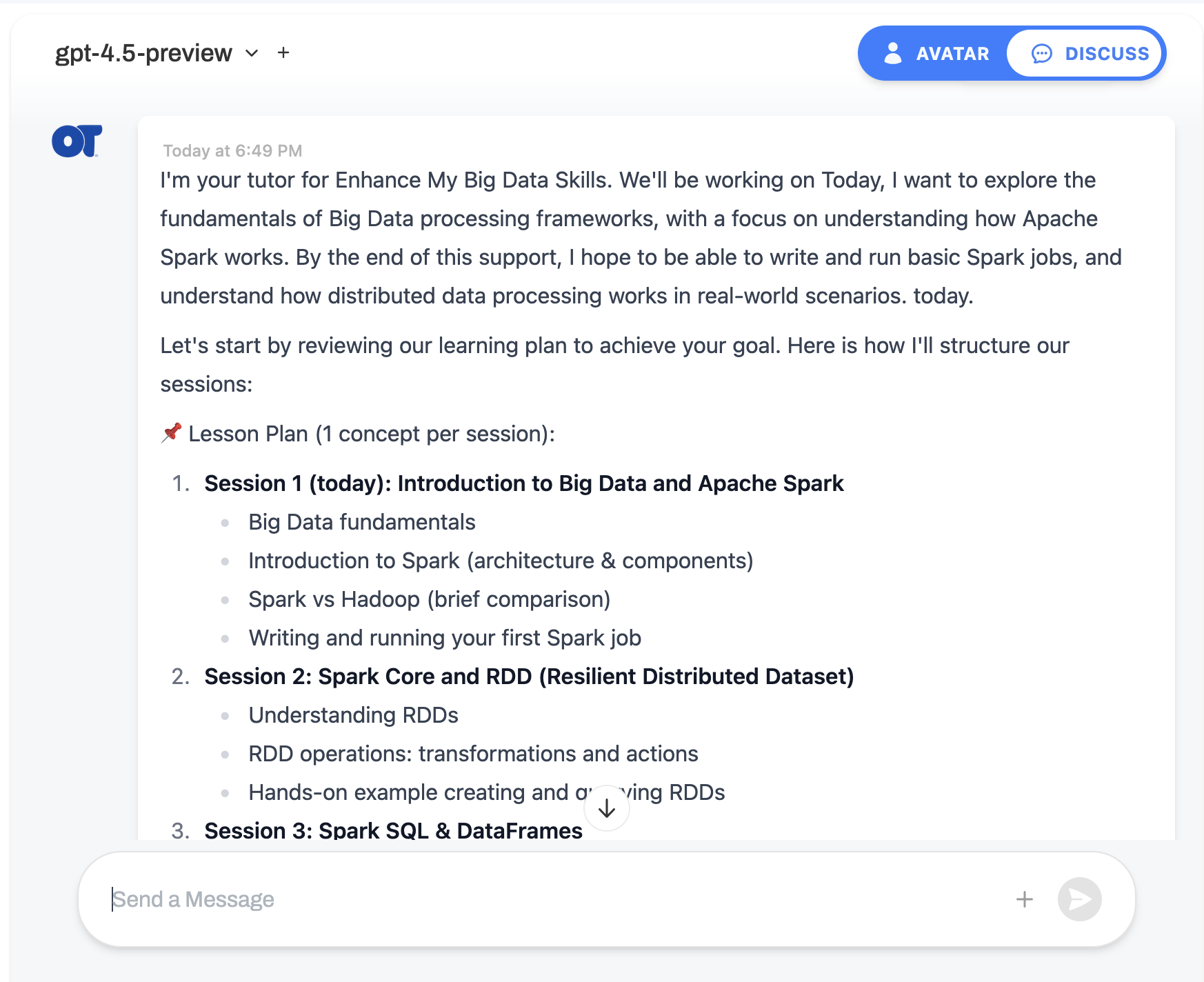}
  \caption{Step 7: The user interacts with a personalized LLM tutor, guided by a dynamically generated learning path based on their submitted materials, goals, level, and preferences. The system adapts explanations, pacing, and interaction style to continuously align with the learner’s evolving needs, making the support deeply targeted and effective.}
  \label{fig:A7}
\end{figure*}

\end{document}